%% file: main.tex
\documentclass{article} 
\usepackage{iclr2027_conference,times}

\usepackage{microtype}
\usepackage{hyperref}
 \hypersetup{
     colorlinks=true,
     linkcolor=citationcolor,
     filecolor=blue,
     citecolor=citationcolor,      
     urlcolor=cyan,
     }
\usepackage{url}            
\usepackage{booktabs}       
\usepackage{amsfonts}       
\usepackage{graphicx}
\usepackage{subfigure}
\usepackage{booktabs} 
\usepackage{amssymb}
\usepackage{pifont}
\usepackage{amsmath}
\usepackage{amsthm}
\usepackage{mathtools}
\usepackage{tikz}
\usepackage{centernot}
\usepackage{caption}
\usepackage{lipsum}  
\usepackage{algorithm}
\usepackage{algorithmic}
\usepackage{etoolbox}\AtBeginEnvironment{algorithmic}{\small} 
\usepackage{wrapfig, blindtext}
\usepackage{needspace}
\usepackage{multirow}
\usepackage{threeparttable}  
\usepackage{enumitem}
\usepackage{listings}
\usepackage{xcolor}

\usepackage{bm} 

\usepackage[most]{tcolorbox}
\usepackage{fontawesome5}

\definecolor{findingpurple}{HTML}{6D66A5}
\definecolor{findinglight}{HTML}{E0E0F3}

\newtcolorbox{findings}{
    enhanced,
    colback=findinglight!40!white,
    colframe=findingpurple!65!white,
    boxrule=1.0pt,
    arc=2.5mm,
    outer arc=2.5mm,
    left=3.5mm,
    right=3.5mm,
    top=1.5mm,
    bottom=1.5mm,
    boxsep=0pt,
    before skip=4pt,
    after skip=4pt,
    before upper={
        \textcolor{findingpurple}{\normalsize\faBookmark}
        \hspace{1.5mm}
        \textbf{Finding.\;\;}
    },
}

\definecolor{findingtan}{HTML}{A98B6F}
\definecolor{findingcream}{HTML}{E8DAC6}
\definecolor{findingink}{HTML}{7A5C3F}

\newtcolorbox{remarks}{
    enhanced,
    colback=findingcream!30!white,
    colframe=findingtan!55!white,
    boxrule=1.0pt,
    arc=2.5mm,
    outer arc=2.5mm,
    left=3.5mm,
    right=3.5mm,
    top=1.5mm,
    bottom=1.5mm,
    boxsep=0pt,
    before skip=4pt,
    after skip=4pt,
    before upper={
        \textcolor{findingink}{\normalsize\faBookmark}
        \hspace{1.5mm}
        \textcolor{findingink}{\textbf{Remark.\;\;}}
    },
}

\definecolor{codeblue}{rgb}{0.25,0.5,0.5}
\definecolor{codekw}{rgb}{0.85, 0.18, 0.50}

\definecolor{codesign}{RGB}{0, 0, 255}
\definecolor{codefunc}{rgb}{0.85, 0.18, 0.50}

\lstdefinelanguage{PythonFuncColor}{
  language=Python,
  keywordstyle=\color{black}\bfseries,
  commentstyle=\color{codeblue},  
  stringstyle=\color{orange},
  showstringspaces=false,
  basicstyle=\ttfamily\small,
}

\input{math.tex}

\input{math_commands.tex}

\renewcommand{\KL}{D_{\texttt{KL}}}

\renewcommand{\t}[1]{\texttt{#1}}

\newcommand{\ttb}[1]{\mathtt{#1}}

\newcommand{\pl}{\purple{\ttb{Long}}}%
\newcommand{\ours}{\textit{LongTake}}%

\definecolor{funcblue}{RGB}{70, 145, 210}
\definecolor{vpurple}{RGB}{115, 75, 170}

\title{LongTake: Learning to Sustain Dynamics \\
in Long-Horizon Video Generation}

\author{
Byoungwoo~Park$^{1}$,~~Jaemoo~Choi$^{2}$,~~Juho~Lee$^{1, \dag}$,~~Yongxin~Chen$^{2, \dag}$ \\
    $^{1}$KAIST, ~~ $^{2}$Georgia Tech, ~~ $^{\dag}$Equal advising\\
}

\iclrfinalcopy 

\begin{document}

\maketitle

\begin{abstract}
World models, game simulators, and long-take video creation
require coherent scene evolution and sustained dynamics
over extended durations.
Autoregressive (AR) video diffusion provides a natural framework
for long-horizon generation,
yet extended rollouts often become near-static or lose visual quality. We hypothesize
that these failures  reflect
the limited guidance provided by short-video supervision
on how ongoing scene dynamics develops over longer durations.
This motivates us to introduce \ours,
a two-stage training pipeline built around
Long-Horizon Teacher Forcing (TF) on curated real long videos.
Long-Horizon TF trains the AR model to predict later frames conditioned on long
ground-truth video prefixes, extending direct supervision beyond the short training horizon. This supervision is designed to help the model sustain dynamics and preserve visual
quality during long-horizon generation. Our central finding is that this training stage strengthens direct initialization for
distribution matching distillation (DMD) under student self-rollout,
without the intermediate few-step distillation stage used in standard pipelines.
Under the same five-second DMD training setup,
our initialization yields substantially higher dynamic degree
than short horizon TF initialization on 30-second rollouts
at comparable aesthetic quality,
and surpasses the evaluated baselines in both measures. Hybrid DMD further reuses this teacher to extend supervision to later frames of the self-rollout while retaining bidirectional joint supervision over the initial window.
On long-horizon self-rollouts, \ours~lies on the Pareto front of dynamic degree and aesthetic quality, and Hybrid DMD attains the highest dynamic degree among evaluated methods at both 30s and 60s.
\end{abstract}

\centerline{\textbf{Project page}~:~ \url{https://bw-park.github.io/LongTake/}}

\section{Introduction}

Recent advances in video generation have enabled visually compelling synthesis
\citep{yang2025cogvideox,kong2024hunyuanvideo,wan2025},
supporting applications in world modeling
\citep{mao2026yume1,sun2026worldplay,hong2025relic,chen2026reworld},
game simulation \citep{tang2025hunyuan,qian2026matrix},
and interactive content creation
\citep{shin2026motionstream,huang2025live,ki2026avatar,xiao2025knot,zhang2026vidu}.
In interactive simulation, generation needs to keep responding to successive actions~\citep{yang2024unisim}
throughout multi-minute sessions \citep{valevski2025gamengen}.
Similarly, a requested long take~\citep{xiong2024lvd} of ``A stylish woman strolls down a bustling Tokyo street." calls for her walk to continue coherently throughout the shot.
A compelling opening is insufficient
if activity fades into near-static frames
or visual quality deteriorates over time.

Autoregressive (AR) video diffusion generates successive chunks
conditioned on a key--value cache (KV-cache) of the video prefix
\citep{yin2025slow,teng2025magi,chen2025skyreels}.
Although many AR models are trained on five-second clips
\citep{huang2025self,zhu2026causal},
these applications require generation beyond the short training horizon (\eg minutes or even an hour).
Recent methods stabilize such generation through context management
\citep{yi2026deep,yesiltepe2026infinity,liu2026rolling,kim2026memrope,mao2026packforcing}
and guidance from the video prefix \citep{song2025historyguided}.
However, generated scenes can still lose visual quality
\citep{cui2026selfforcing,li2026rolling} or settle into near-static or repetitive
states~\citep{henschel2025streamingt2v,lu2026reward,dalva2026adastate}. Such stagnation is easy to miss, as consistency metrics can favor low-dynamic videos
even when the requested dynamics are no longer sustained
\citep{liao2024devil,huang2024vbench}.

These failures motivate learning long-horizon generation from real long videos,
which show how a scene keeps evolving while remaining visually coherent
over tens of seconds~\citep{xiong2024lvd}.
Prior work shows that extending the temporal span of training sequences
can improve the modeling of temporal dynamics,
since sequences shorter than an effect provide little training signal for it~\citep{brooks2022generating}.
Training on short video clips thus provides no direct supervision
for effects that unfold over tens of seconds
and leaves the model to generalize beyond its training horizon.
Pairing long video prefixes with the frames that follow them
instead extends supervision to later stages of the same evolving scene,
where long-horizon generation takes place.
This leads to the following question.

\graybox{%
\
\begin{center}
\textit{Can supervision from real long videos help sustain coherent dynamics while \\ preserving visual quality as the scene evolves during long-horizon autoregressive rollout?}
\end{center}
}

\begin{figure*}[t]
\centering
\includegraphics[width=\linewidth]
{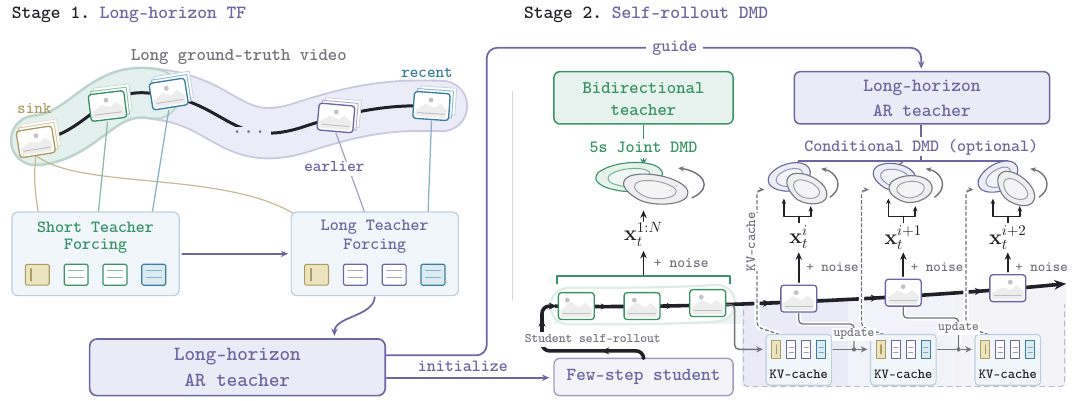}
\caption{
\textbf{LongTake learns long-horizon generation
through a two-stage training pipeline.}
(Left) \textit{Stage 1:} Long-Horizon TF learns from real long videos
and provides a strong direct initialization for self-rollout DMD.
(Right) \textit{Stage 2:} Self-rollout DMD uses this initialization.
Hybrid DMD further adds conditional supervision of later frames
from the AR teacher,
while retaining joint supervision
from the frozen bidirectional teacher
over the initial window of the student rollout.
}
\vspace{-3mm}
\label{fig:intro_conceptual}
\end{figure*}

We introduce~\ours, a two-stage training pipeline
built around Long-Horizon Teacher Forcing (TF) on curated real long videos as described
in~\Cref{fig:intro_conceptual}.
Here, Long-Horizon TF constructs the conditioning KV-cache from long ground-truth video
prefixes
and directly supervises later frames beyond the short training horizon. Our central finding is that the resulting model
serves as a strong direct initialization
for self-rollout distribution matching distillation (DMD)
\citep{zhuang2026self}, supporting effective two-stage training without a separate
few-step initialization stage~\citep{zhu2026causal,zhao2026causal}. Under the same 5s
DMD setting~\citep{zhuang2026self}, initializing from the Long-Horizon TF model instead
of a standard short-clip TF model raises the VBench dynamic
degree~\citep{huang2024vbench} of 30s rollouts while preserving comparable aesthetic
quality and also surpasses the three-stage pipeline in both measures.  Beyond initialization, extending DMD to later frames
relies on teacher scores under longer video prefixes,
where a 5s DMD teacher may provide limited guidance. Hybrid DMD therefore reuses the Long-Horizon TF model
to provide conditional scores for later frames
of the student self-rollout. On 30s and 60s rollouts, both LongTake
variants lie on the Pareto front
of dynamic degree against aesthetic quality and against non-motion VBench quality,
as the only methods on either front above 90 in dynamic degree.

Our contributions are summarized below.
\begin{itemize}[leftmargin=12pt]
\item We curate real long videos and introduce Long-Horizon TF to supervise later frames
from long video prefixes beyond the short training horizon.
\item We show that Long-Horizon TF on curated real long videos strengthens initialization
for the same 5s self-rollout DMD,
yielding stronger long-horizon dynamics at comparable visual quality
without the separate few-step initialization stage
of the standard pipeline.

\item We introduce Hybrid DMD, which reuses the Long-Horizon TF model
as a teacher for later frames of the student self-rollout
and attains the highest dynamic degree.

\item On 30s and 60s rollouts, \ours~lies on the Pareto front of dynamic degree and aesthetic quality
with dynamic degree above 90,
whereas baselines with higher aesthetic quality stay at or below 58.
\end{itemize}

\section{Related Work}
\label{sec:related-work}

\paragraph{Long-video supervision and AR initialization.}
Real long videos provide direct supervision
for scene evolution beyond short clips.
LVD-2M~\citep{xiong2024lvd}
and Presto~\citep{yan2025long}
emphasize dynamic long-take videos,
while Long Context Tuning~\citep{guo2025long}
learns dependencies across extended scenes.
For AR initialization,
Causal Forcing~\citep{zhu2026causal}
and Causal Forcing++~\citep{zhao2026causal}
use separate causal ODE and consistency distillation stages
before self-rollout DMD.
SGF~\citep{zhuang2026self}
also evaluates direct initialization from a TF-trained model.
Building on this approach, LongTake studies how supervision
from curated real long videos strengthens direct initialization.
To this end, Long-Horizon TF supervises later frames
under KV-cache states constructed from long video prefixes.
Under the same 5s joint DMD,
this initialization yields stronger long-horizon dynamics
while preserving comparable visual quality.

\paragraph{Long-horizon self-rollout distillation}
Self Forcing~\citep{huang2025self} trains under student self-rollout to reduce the \emph{training--inference gap}.
Self-Forcing++~\citep{cui2026selfforcing} applies DMD to short windows from extended student rollouts,
while \citet{cai2026mode} combine long-video supervision with sliding-window distribution matching.
Context Forcing~\citep{chen2026context} also trains a long-context teacher on real videos
and jointly matches a target window under the preceding student self-rollout. Context-Matched Distillation~\citep{bandyopadhyay2026context}
initializes the student from its causal teacher
and scores each chunk under its own student-generated prefix.
Hybrid DMD scores each next chunk in the same way
but reuses the Long-Horizon AR teacher
and retains bidirectional joint supervision over the initial window.
Further comparisons appear in Appendix~\ref{appendix:Related work}.

\section{Preliminaries}
\label{sec:preliminaries}

\begin{wrapfigure}[17]{r}
{0.297\linewidth}
\vspace{-5mm}
\centering
    \includegraphics[width=1.\linewidth]{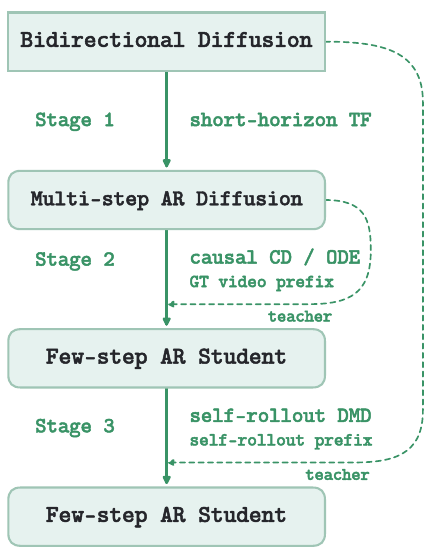}
\caption{Three-stage pipeline for few-step AR video diffusion~\citep{zheng2026causal}.}
\label{fig:standard_pipeline}
\end{wrapfigure}

In this section, we review the three-stage training pipeline
commonly used for few-step AR video diffusion
\citep{zhu2026causal,zhao2026causal,zheng2026causal}.
As summarized in~\Cref{fig:standard_pipeline},
AR diffusion training is followed by a separate
few-step initialization stage and self-rollout DMD.
We use this pipeline as a reference to examine
how Long-Horizon TF strengthens direct initialization for DMD.

\paragraph{Stage 1: AR diffusion training}
AR video diffusion generates a video as a sequence of chunks\footnote{We use chunk-level notation throughout the paper, with frame-wise AR as the single-frame chunk case.}, with the AR factorization
\[
    p(\bx)= \textstyle{\prod_{i=1}^{N}} p^{\ttb{AR}}(\bx^i | \bx^{<i}),
\]
where $\bx^i$ denotes a chunk of consecutive latent frames. A common starting point for AR diffusion distillation is to fine-tune a pretrained bidirectional diffusion model with teacher forcing (TF)~\citep{jin2025pyramidal}. Each noisy chunk $\bx_{t}^i$ at a sampled noise level $t$ is conditioned on the keys and values computed by the AR model from the preceding clean ground-truth video prefix $\bx^{<i}_{\texttt{gt}}$. This gives the standard AR flow-matching objective for short-horizon TF
\[
    \phi^{\star} = \argmin_{\phi} \bbE_{i \in [1, N], t_i,\varepsilon_i, \bx_{\ttb{gt}} \sim \calD}
    \left[w(t_i)\left\|v_\phi^{\ttb{AR}}(\bx_{t_i}^i, \bx_{\ttb{gt}}^{<i}, \varnothing,t_i)-(\varepsilon_i - \bx_{\ttb{gt}}^i)\right\|^2\right],
    \label{eq:ar-flow-matching}
\]
where $\bx_{t_i}^i = (1-t_i) \bx_{\ttb{gt}}^i + t_i \varepsilon_i$ and $\varepsilon_i \sim \calN(0, I)$. Standard TF in~\eqref{eq:ar-flow-matching} is applied to a short ground-truth clip $\bx_{\t{gt}} \sim \calD_{\ttb{short}}$ containing at most $N$ frames ($\eg$ $21$ latent frames). The model receives the clean video prefix $\bx^{<i}_{\ttb{gt}}$ and an empty initial KV-cache $\varnothing$. Using causal masking, the AR model evaluates losses for all target chunks in parallel in one forward pass.

\paragraph{Stage 2: Few-step initialization} The multi-step AR diffusion model $v_{\phi^{\star}}^{\ttb{AR}}$ from~\eqref{eq:ar-flow-matching} can be further converted into a few-step generator $v^{\ttb{AR}}_{\theta^{\star}}$ before self-rollout training. Existing pipelines perform this initialization using ODE distillation~\citep{zhu2026causal}, consistency distillation~\citep{zhao2026causal,zheng2026causal}, where the AR teacher and student use the same ground-truth video prefix from $\bx_{\ttb{gt}} \sim \calD_{\ttb{short}}$. This stage provides a few-step initialization for the subsequent self-rollout distribution-matching stage using student-generated videos.

\paragraph{Stage 3: Distribution matching distillation}
\label{sec:distribution-matching}
Few-step initialization uses ground-truth video prefixes $\bx_{\ttb{gt}}^{<i}$. At inference, the conditioning KV-cache is instead updated from student-generated chunks. Self-rollout training~\citep{huang2025self} reduces this gap by generating according to the factorization
\[
    q^{\theta}(\bx_{\ttb{self}}) = \textstyle{\prod_{i=1}^{N}} q^{\theta} \left(\bx_{\ttb{self}}^i | \calC_{\ttb{cache}}(\bx_{\ttb{self}}^{<i})\right),
    \label{eq:student}
\]
where $\calC_{\ttb{cache}}$ maps a video prefix to a bounded per-layer KV-cache that retains keys and values from preceding chunks. This KV-cache is updated after each generated chunk according to the cache policy. Distribution matching distillation (DMD)~\citep{yin2024one, huang2025self} uses a frozen bidirectional teacher $p_t^{\star}$ to supervise the joint distribution $q_t^{\theta}$ of the student rollout through:
\[
    \calJ_{\ttb{DMD}}(\theta)= \bbE_{t, \bx_{\ttb{self}} \sim q^{\theta}} \left[w(t) \KL \left(q_t^\theta(\bx_t) | p_t^\star(\bx_t)\right)\right].
    \label{eq:joint-dmd}
\]
Hence, DMD in~\eqref{eq:joint-dmd} trains the generator on self-rollouts $\bx_{\ttb{self}}$, exposing it to KV-cache constructed from its own generated chunks and reducing the \textit{training--inference gap}~\citep{huang2025self}.

\section{Training for Long-Horizon Video Generation}

In this section, we first introduce Long-Horizon TF to learn long-horizon generation
from real long videos.
Our central finding is that the resulting initialization
substantially improves long-horizon generation
under the same short-horizon self-rollout DMD procedure,
enabling effective two-stage training.
Hybrid DMD further uses this AR teacher
to supervise later frames of the student rollout.

\subsection{Long-Horizon Supervision for Autoregressive Generation}
\label{sec:horizon-gap}

\begin{wrapfigure}[13]{r}
{0.35\linewidth}
\vspace{-5mm}
\centering
    \includegraphics[width=1.\linewidth]{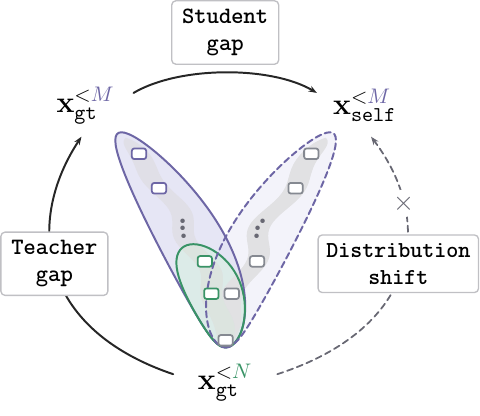}
\caption{
\textbf{Teacher and student gaps in KV-cache states.}
}
\label{fig:kv_cache_long_dataset}
\end{wrapfigure}

Let $\green{N}$ denote the short-video training horizon
and $\purple{M}\gg\green{N}$ the target rollout horizon.
The standard pipeline in~\Cref{fig:standard_pipeline}
trains the AR diffusion model and performs self-rollout DMD
over $\green{N}$ (\eg $21$ latent frames).
During AR training, the ground-truth video prefix
$\bx_{\ttb{gt}}^{<i}$ is encoded into the conditioning KV-cache,
and learning from short clips directly supervises
$p_{\phi}^{\ttb{AR}}(\bx^i \mid \bx_{\ttb{gt}}^{<i})$
only for $i<\green{N}$.
Long-horizon generation, however, requires the model
to continue from longer video prefixes
at positions $i \geq \green{N}$.
Because short-clip training provides no direct supervision
for prediction at these later positions,
the model relies on generalization to continue
beyond the short-video training horizon through the AR self-rollout.


\paragraph{Teacher gap}
The \textit{teacher gap} concerns the shift from KV-cache states
constructed from ground-truth videos within $\green{N}$
to those encountered toward $\purple{M}$.
Real long videos provide the missing supervision
by pairing later conditioning KV-cache states
with corresponding ground-truth target frames.
Long-Horizon TF trains the AR teacher on these pairs
to generate later frames, as detailed in~\Cref{sec:long-teacher}.

\paragraph{Student gap}
Long-Horizon TF expands teacher supervision
to KV-cache states constructed from ground-truth videos
throughout $\purple{M} \gg \green{N}$.
At inference, however, the conditioning KV-cache is
constructed from student self-rollout
$\bx_{\ttb{self}}^{<i}$.
These states can differ from those constructed
from the ground-truth video prefixes $\bx_{\ttb{gt}}^{<i}$ used in TF.
We refer to this familiar
\textit{training--inference gap}~\citep{huang2025self}
as the \textit{student gap}.
Self-rollout DMD addresses this mismatch
by training under student-generated video prefixes.
\Cref{fig:kv_cache_long_dataset} summarizes the teacher gap and student gap,
which concern the training horizon and the transition to student self-rollout,
respectively.

\begin{figure*}[t]
\centering
\begin{minipage}{0.95\linewidth}
    \centering
    \includegraphics[width=0.95\linewidth]{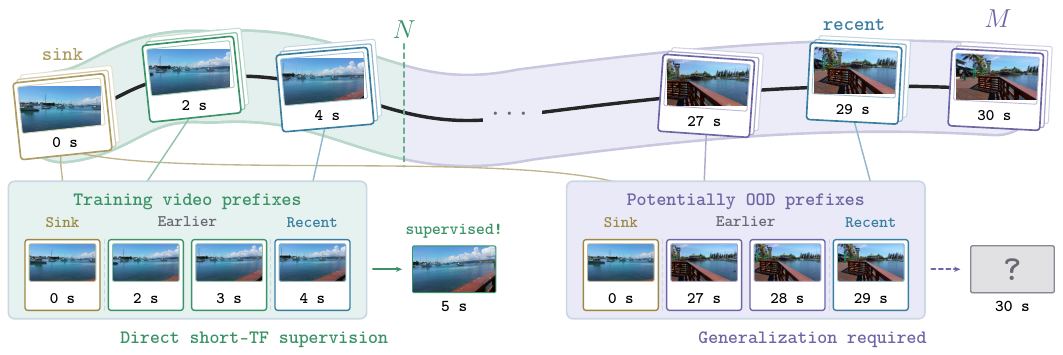}
\end{minipage}
\caption{
\textbf{Teacher gap with a bounded KV-cache.}
With sink retention,
the temporal separation between sink frames and recent frames
increases as the video prefix grows.
Short-horizon TF directly supervises prediction
only within $\green{N}$,
while generating later frames toward $\purple{M}$
relies on generalization.
}
\label{fig:kv_cache_evolving_states}
\vspace{-3mm}
\end{figure*}

\subsection{Long-Horizon Teacher Forcing}
\label{sec:long-teacher}

We extend teacher forcing to prediction positions beyond $\green{N}$
using real long videos $\bx_{\ttb{gt}}$
from the curated dataset $\calD_{\pl}$.
Long-Horizon TF constructs the conditioning KV-cache
by applying the chosen cache update rule
to longer ground-truth video prefixes.
Supervision on the later target frames teaches the AR model to predict them from the temporal context preserved as the same scene evolves over extended durations, which targets the \textit{teacher gap}.

\paragraph{Long-Horizon TF}
Given a ground-truth video $\bx_{\ttb{gt}}\sim\calD_{\pl}$,
we sample a valid split $s$
and apply TF to a target window of length $\green{N}$ beginning at $s$.
The conditioning KV-cache at the split is
$\ttb{c}_{\ttb{gt}}^s
:=\calC_{\ttb{cache}}(\bx_{\ttb{gt}}^{<s})$,
where the chosen cache update rule determines
which information from the video prefix is preserved.
With sink retention~\citep{yang2026longlive},
the initial frames remain in the conditioning KV-cache
while the recent frames are updated as the video prefix grows.
Sampling longer video prefixes therefore provides supervision over larger temporal separations between the retained early and recent frames, covering how later frames relate to this context. For each target chunk at $i\in[s,s+\green{N})$,
the model uses this cache and preceding ground-truth frames
$\bx_{\ttb{gt}}^{s<i}$ within the target window,
giving the flow-matching objective defined as follows:

\graybox{
\[
    \phi^\star
    =
    \argmin_{\phi}
    \bbE_{
        s, i\in[s,s+\green{N}),
        t_i,\varepsilon_i,
        \bx_{\ttb{gt}}\sim\calD_{\pl}
    }
    \left[
    w(t_i)
    \norm{
    v_\phi^{\ttb{AR}}
    \left(
        \bx_{t_i}^i,
        \bx_{\ttb{gt}}^{s<i},
        \purple{\ttb{c}_{\ttb{gt}}^s},
        t_i
    \right)
    -
    (\varepsilon_i-\bx_{\ttb{gt}}^i)
    }^2
    \right].
    \label{eq:long-teacher-training}
\]
}

When $s=0$, the conditioning KV-cache is empty,
$\ie \ttb{c}_{\ttb{gt}}^s=\varnothing$,
recovering standard TF in~\eqref{eq:ar-flow-matching}.

\begin{wrapfigure}[16]{r}
{0.45\linewidth}
\vspace{-10mm}
\centering
\begin{minipage}{0.97\linewidth}
\begin{algorithm}[H]
\caption{Long-Horizon TF \\
{\scriptsize{Note: Details of parallel computation are in Appendix~\ref{appendix:parallel cache prefill}.}}
}
\label{alg:long-teacher}
\vspace{-2.5mm}
\begin{lstlisting}[
language=PythonFuncColor,
mathescape=true,
backgroundcolor=\color{orange!3},
basicstyle=\fontsize{6.5pt}{7.2pt}\ttfamily\selectfont,
emph={[1]v, M},
emphstyle={[1]\color{vpurple}\bfseries},
emph={[2]sample,len,randint,randn_like,sample_timestep,mse_loss,update},
emphstyle={[2]\color{funcblue}},
emph={[3]c, N, parallel_prefill},
emphstyle={[3]\color{mgreen}\bfseries},
]
# v - autoregressive flow model
# C - bounded KV-cache
# N - short TF horizon
while True:
    x = sample($\mathcal{D}_{\color{vpurple}\bm{\mathtt{Long}}}$)
    M = len(x)
    s = randint(0, M - N)

    # Pass 1: Parallel cache prefill
    prefix = x[:s]
    c = parallel_prefill(prefix)

    # Pass 2: Teacher-forcing
    x_tar = x[s:s+N]
    t = sample_timestep()
    eps = randn_like(x_tar)
    x_t = (1 - t) * x_tar + t * eps
    v_tar = eps - x_tar
    v_hat = v(x_t, x_tar, c, t)
    
    loss = mse_loss(v_hat, v_tar)
    update(v, loss)
return v
\end{lstlisting}
\vspace{-2.5mm}
\end{algorithm}
\end{minipage}
\end{wrapfigure}

\paragraph{Parallel cache computation}
The entire conditioning KV-cache can be constructed in parallel
for both sink retention with FIFO eviction~\citep{yang2026longlive}.
Specifically, we compute prefix \t{Q/K/V}s
in parallel with causal masking
and select retained entries according to their temporal positions.
For the EMA extension~\citep{lu2026reward},
a parallel prefix scan~\citep{blelloch1990prefix}
incorporates the accumulated contribution of evicted entries into the sink.
Both variants therefore produce $\ttb{c}_{\ttb{gt}}^s$
for each layer within the same parallel cache-construction pass.
Then, next pass computes TF losses over the fixed target window as in~\eqref{eq:ar-flow-matching}.
The clean stream supplies preceding ground-truth frames,
while the noisy stream predicts target velocities.
For $s>0$, Long-Horizon TF thus separates cache construction
from target prediction into \textit{two forward passes},
as summarized in~\Cref{alg:long-teacher}.

\begin{figure*}[t]
\begin{minipage}{1.\linewidth}
    \includegraphics[width=1.\linewidth]{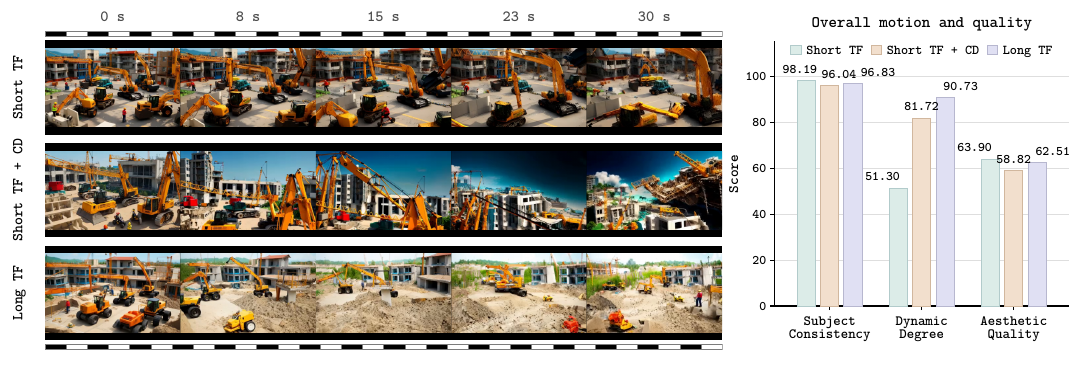}
\end{minipage}
\caption{
\textbf{Long-Horizon TF enables effective two-stage training.}
All models use $\green{5}$s self-rollout DMD
and are evaluated on $\purple{30}$s rollouts.
Long TF initialization improves long-horizon dynamics
over Short TF while maintaining comparable visual quality,
showing that stronger TF initialization supports
effective long-horizon generation
without a separate CD stage.
}
\label{fig:video_difference_with_scores}
\vspace{-4mm}
\end{figure*}

\paragraph{Effect of Long-Horizon TF initialization}
Following SGF~\citep{zhuang2026self},
we initialize self-rollout DMD directly from a TF-trained model
while retaining the same DMD procedure
and its short horizon $\green{N}$.
In~\Cref{fig:video_difference_with_scores},
compared with short-horizon TF initialization, causal CD initialization produces stronger dynamics but lower aesthetic quality. Long-Horizon TF initialization improves dynamics while retaining comparable aesthetic quality to short-horizon TF initialization.

\begin{findings}
\textit{Long-Horizon TF on curated real long videos strengthens direct initialization for self-rollout DMD.
Under the same DMD procedure,
the resulting student achieves stronger long-horizon dynamics
with comparable visual quality,
without a separate few-step initialization.}
\end{findings}

\subsection{Extending Supervision with Hybrid DMD}
\label{sec:conditional-dmd}

We next extend the benefit of Long-Horizon TF
by reusing the resulting AR teacher
to supervise later frames under student self-rollout.
Hybrid DMD adds this conditional supervision
while retaining joint supervision over the initial window
from the frozen bidirectional teacher.

\begin{wrapfigure}[14]{r}{0.41\linewidth}
\centering
\begin{minipage}{0.41\textwidth}
\vspace{-10mm}
\begin{algorithm}[H]
\caption{Hybrid DMD \\
{\scriptsize{Note: Details of Hybrid DMD are in Appendix~\ref{appendix:method details}.}}
}
\label{alg:conditional-forcing}
\vspace{-2mm}
\begin{lstlisting}[
language=PythonFuncColor,
mathescape=true,
backgroundcolor=\color{orange!3},
basicstyle=\fontsize{6.5pt}{7.2pt}\ttfamily\selectfont,
emph={[1]G, cond_DMD},
emphstyle={[1]\color{vpurple}\bfseries},
emph={[2]self_rollout,update},
emphstyle={[2]\color{funcblue}},
emph={[3]C, joint_DMD},
emphstyle={[3]\color{mgreen}\bfseries},
]
# G - few-step AR generator
# C - bounded KV-cache policy
# N - initial joint-DMD horizon
while True:
    # Student self-rollout
    x, C = self_rollout(G)
    
    # Hybrid DMD
    l_jnt = joint_DMD(x[:N],$\varnothing$,teacher=$\green{p^{\ttb{Bi}}}$)

    l_cnd = 0
    for i in range(N, len(x)):
        l_cnd += cond_DMD(x[i],C[i], teacher=$\purple{p^{\ttb{AR}}}$)
    l_cnd /= len(x[N:])

    loss = l_jnt + $\lambda$ * l_cnd
    update(G, loss)
return G
\end{lstlisting}
\vspace{-2mm}
\end{algorithm}
\end{minipage}
\end{wrapfigure}

\paragraph{Extending supervision}
Self-rollout DMD~\citep{huang2025self}
trains the student using its own generated frames,
but supervision over the initial window of length $\green{N}$
provides no direct training signal for later frames.
Long-Horizon TF in~\eqref{eq:long-teacher-training}
trains the AR teacher $p_{\phi^\star}^{\ttb{AR}}$
to generate frames beyond $\green{N}$.
For each target chunk beyond the initial window, we construct
its conditioning KV-cache
$\ttb{c}_{\ttb{self}}^i
:=\calC_{\ttb{cache}}(\bx_{\ttb{self}}^{<i})$
from the student-generated video prefix.
The teacher then provides a conditional target
$p^{\ttb{AR}}_{\phi^{\star}}(\cdot \mid \ttb{c}_{\ttb{self}}^i)$
for that chunk.
Whereas Context Forcing~\citep{chen2026context}
jointly scores a target window,
our conditional DMD scores each chunk
using the KV-cache updated from preceding student-generated frames.
Appendix~\ref{sec:Comparison with existing work.}
compares the two scoring objectives and their conditioning video prefixes
at each target chunk.

\paragraph{Hybrid DMD}
We retain the pretrained bidirectional teacher
$p^{\ttb{Bi}}$ for joint supervision
over the initial window of length $\green{N}$
and use the Long-Horizon AR teacher
to supervise later frames.
Together, they define the hybrid teacher distribution
for student self-rollout over the horizon $\purple{M}$ as
\begin{equation}
    p^\star(\bx_{\ttb{self}}^{0:\purple{M}})
    =
    p^{\ttb{Bi}}
    (\bx_{\ttb{self}}^{0:\green{N}})
    \textstyle{\prod_{i=\green{N}}^{\purple{M}-1}}
    p_{\phi^\star}^{\ttb{AR}}
    \left(
        \bx_{\ttb{self}}^i
        \mid \ttb{c}_{\ttb{self}}^i
    \right).
    \label{eq:hybrid-teacher-distribution}
\end{equation}
We apply joint DMD over the initial window
and conditional DMD beyond that window,
with $\lambda>0$ controlling the contribution
of conditional supervision relative to joint supervision.
Combining these terms gives the Hybrid DMD objective
for joint and conditional distribution matching over $\purple{M}$ as

\graybox{
\
\begin{equation}
\begin{aligned}
    \calJ_{\ttb{Hybrid-DMD}}(\theta)
    ={}&
    \KL\left(
        q^\theta
        (\bx_{\ttb{self}}^{0:\green{N}})
        \,\middle\|\,
        p^{\ttb{Bi}}
        (\bx_{\ttb{self}}^{0:\green{N}})
    \right)
    \\
    &+
    \lambda
    \bbE_{i \in [\green{N}, \purple{M})}
    \bbE_{\bx_{\ttb{self}}^{<i}\sim q^\theta}
    \left[
        \KL\left(
            q^\theta
            (\bx_{\ttb{self}}^i
            \mid \ttb{c}_{\ttb{self}}^i)
            \,\middle\|\,
            p_{\phi^\star}^{\ttb{AR}}
            (\bx_{\ttb{self}}^i
            \mid \ttb{c}_{\ttb{self}}^i)
        \right)
    \right].
\end{aligned}
\label{eq:hybrid-dmd}
\end{equation}
}

Here, the expectation averages conditional matching over student-generated
video prefixes $\bx_{\ttb{self}}^{<i}$ and the teacher and student construct their conditioning KV-cache
from the same prefix while retaining their respective representations.
Both teachers remain frozen throughout the joint and conditional supervision
of student-generated frames,
as summarized in~\Cref{alg:conditional-forcing}. 
\Needspace{6\baselineskip}
\begin{wrapfigure}[16]{r}
{0.31\linewidth}
\begin{minipage}[t]{0.31\textwidth}
\vspace{-6mm}
\centering
    \includegraphics[width=1.\linewidth]{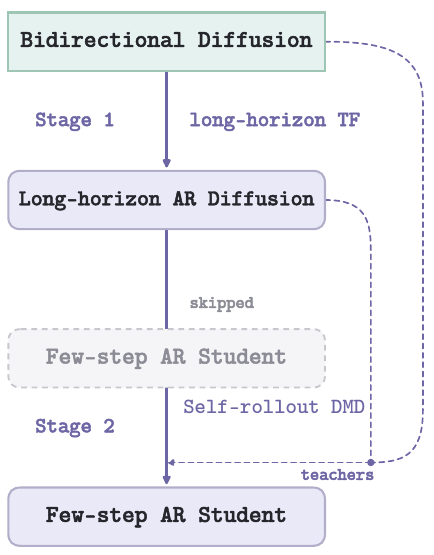}
\caption{\textbf{Two-stage training}}
\label{fig:our_pipeline}
\end{minipage}
\end{wrapfigure}

\subsection{Proposed training pipeline}
We combine Long-Horizon TF and self-rollout DMD
into the two-stage training pipeline shown in~\Cref{fig:our_pipeline}.
The first stage trains an AR diffusion model
on the real long-video dataset $\calD_{\pl}$
using Long-Horizon TF in~\eqref{eq:long-teacher-training}.
The resulting weights directly initialize the student
for the second stage of self-rollout DMD,
without a separate few-step initialization stage.
We begin with short-horizon joint DMD,
where the frozen bidirectional teacher
supervises the joint distribution of student self-rollouts
of length $\green{N}$.
This completes the standard two-stage \ours~pipeline.
As an extension within the second stage,
Hybrid DMD continues training on longer student rollouts.
The bidirectional teacher retains joint supervision
over the initial window of length $\green{N}$,
while the TF-trained AR model serves as the frozen
Long-Horizon AR teacher for conditional supervision of later frames.
The AR teacher conditions on the preceding student-generated video prefix,
and both teachers remain frozen as the student is updated.
Appendix~\ref{appendix:method details}
details the student updates
and the gradient replay procedure used to compute them.

\section{Experiments}

\paragraph{Implementation details}
We use $\ttb{Wan2.1}$-$\ttb{T2V}$-$\ttb{1.3B}$~\citep{wan2025}
to generate videos at $832\times480$ and 16 FPS,
with three latent frames per AR chunk and four denoising steps.
Long-Horizon TF uses 43K video--text pairs,
including OpenVidHD clips~\citep{nan2025openvidm},
and 3,000 updates to supervise generation up to 30s.
The resulting model directly initializes self-rollout DMD
using SGF~\citep{zhuang2026self}
with extended VidProM prompts~\citep{wang2024vidprom}.
LongTake uses 1,200 joint DMD iterations on 21-latent-frame rollouts.
The Hybrid DMD variant uses 800 joint DMD iterations
followed by 400 Hybrid DMD iterations on 42-latent-frame rollouts
with $\lambda=0.2$.
Both variants use $\ttb{Wan2.1}$-$\ttb{T2V}$-$\ttb{14B}$
as the bidirectional teacher and a fixed 12-latent-frame
student attention window.
Each variant uses its final EMA checkpoint for both evaluation durations.
Appendix~\ref{appendix:experimental details} provides
checkpoint sources and complete training configurations for both stages.

\paragraph{Baselines and evaluation}
We compare publicly released $\ttb{1.3B}$ checkpoints of
Self Forcing~\citep{huang2025self},
Causal Forcing~\citep{zhu2026causal},
LongLive~\citep{yang2026longlive},
Reward Forcing~\citep{lu2026reward},
MemRoPE~\citep{kim2026memrope},
Rolling Forcing~\citep{liu2026rolling},
SGF~\citep{zhuang2026self},
and Context Forcing~\citep{chen2026context}
in~\Cref{tab:main_results}.
We evaluate 30s and 60s generation on
128 extended MovieGen prompts~\citep{polyak2024movie}
and 251 VBench-Long prompts~\citep{huang2025vbench++}, respectively.
Quality uses the standard VBench~\citep{huang2024vbench} normalization and weights
across seven video-quality dimensions.
We separately report
\texttt{\textbf{$\Delta$IQ}}~\citep{liu2026rolling}
and \textbf{\texttt{VLM}} exposure~\citep{kim2026memrope}
to assess long-horizon visual stability.

\begin{figure*}[t]
\begin{minipage}{1.\linewidth}
    \includegraphics[width=1.\linewidth]{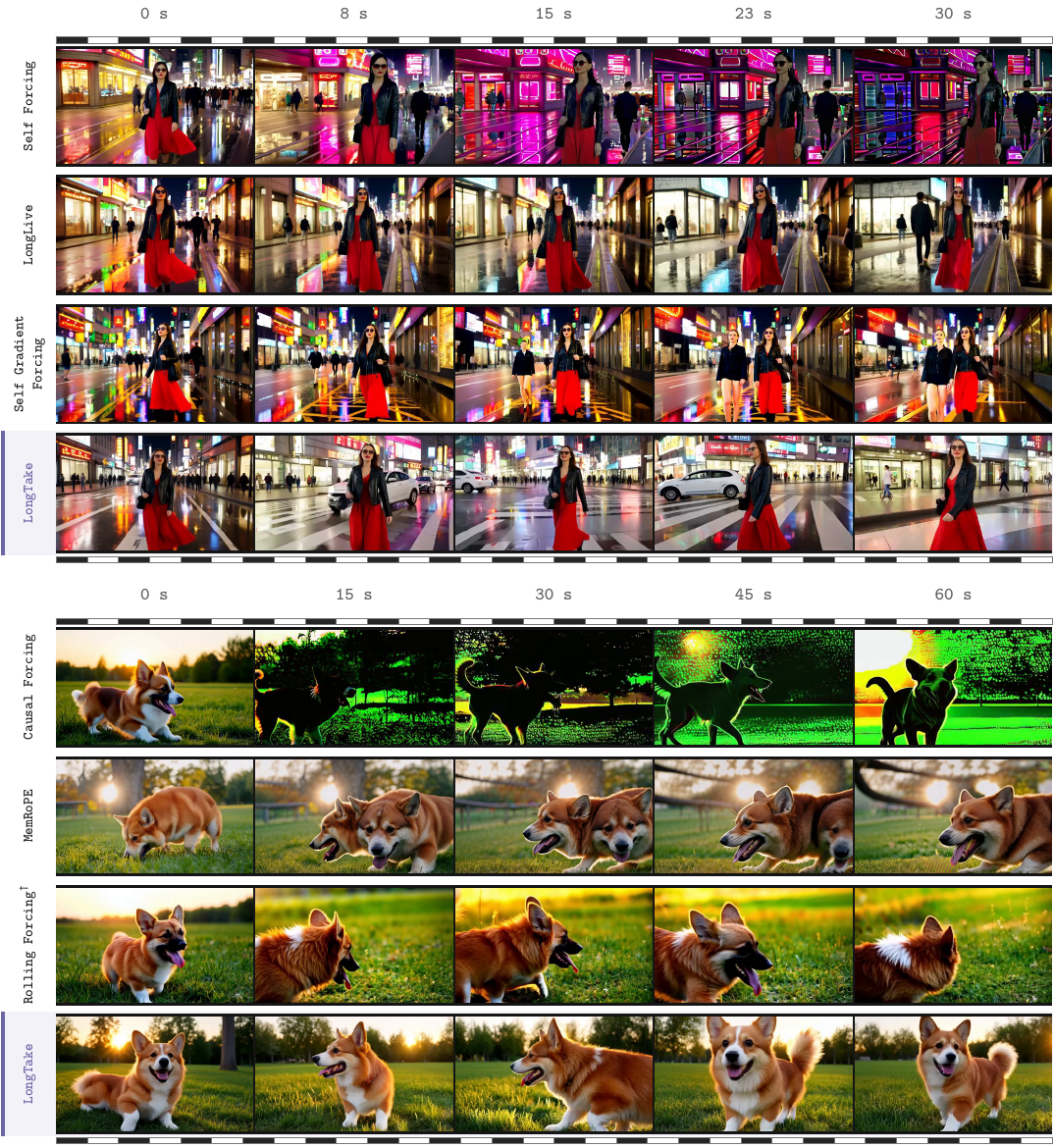}
\end{minipage}
\caption{
\textbf{Long-horizon generation with sustained dynamics.}
Qualitative comparisons over 30s (top) and 60s (bottom).
Baseline rollouts exhibit appearance degradation
or limited scene progression.
LongTake with Hybrid DMD sustains scene evolution
with coherent appearance in later frames.
}
\label{fig:qualitative_30s_60s}
\vspace{-6mm}
\end{figure*}

\paragraph{Qualitative results}
\Cref{fig:qualitative_30s_60s} compares baselines with LongTake using Hybrid DMD.
In Tokyo, the three methods with the lowest dynamic degree at 30s
keep the woman in nearly the same pose and position from 8s to 30s,
so the requested walk stalls even where frames stay sharp,
whereas LongTake keeps her walking as the scene changes.
At 60s, beyond the 30s supervision horizon, LongTake preserves the dog's appearance,
while Causal Forcing develops color artifacts, MemRoPE duplicates the subject,
and Rolling Forcing drifts into close-ups despite high dynamic degree.

\begin{wrapfigure}[15]{r}
{0.35\linewidth}
\begin{minipage}[t]{0.35\textwidth}
\centering
    \includegraphics[width=1.\linewidth]{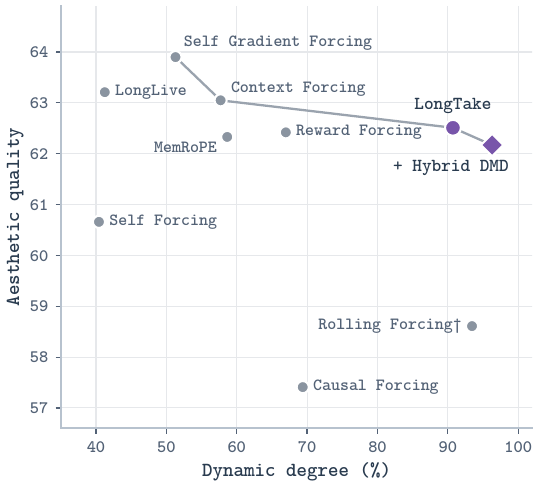}
\caption{\textbf{Motion-quality trade-off at 30s.}
Gray line: Pareto front.
}
\label{fig:dynamics_aesthetics}
\end{minipage}
\end{wrapfigure}

\paragraph{Quantitative results}
As shown in~\Cref{fig:dynamics_aesthetics}, the baselines trade motion for quality. Every baseline with higher aesthetic quality than LongTake reaches a dynamic degree of at most $58$, while the only one above $90$, Rolling Forcing, trails LongTake by $3.9$ aesthetic score. LongTake lies on the Pareto front at both durations (see Appendix~\ref{appendix:motion-quality trade-off}) and, relative to other methods on the front, gains $33$-$42$ scores of dynamic degree for under two aesthetic scores. Reward Forcing, the strongest baseline in Quality at $60$s, matches LongTake in aesthetic quality with $14.5$ fewer score of dynamic degree. In~\Cref{tab:main_results}, LongTake with only $5$s joint DMD also exceeds every baseline in Quality at both durations. Hybrid DMD achieves the highest dynamic degree at both durations and the highest Quality at $30$s, while joint DMD retains higher Quality at $60$s, so the preferred variant depends on the target duration.

\begin{table*}[t]
\centering
\caption{
\textbf{Evaluation on 30s and 60s AR rollouts.}
\texttt{\textbf{Quality}} aggregates seven video-quality dimensions
with standard VBench normalization and weights.
\texttt{\textbf{$\Delta$IQ}} and \texttt{\textbf{VLM}} assess visual stability.
\best{Best} and \secondbest{second-best} results are highlighted for each duration.
LongTake uses Long-Horizon TF followed by 5s joint DMD.
Full results appear in~\Cref{table:full_vbench_table}.
$\dagger$ checkpoint from~\citep{zhu2026causal}.
}
\label{tab:main_results}
\resizebox{\textwidth}{!}{%
\renewcommand{\arraystretch}{1.2}
\setlength{\tabcolsep}{1pt}
\begin{tabular}{@{}l cccc cc|cccc cc@{}}
\toprule
& \multicolumn{6}{c}{\texttt{\textbf{30s}}}
& \multicolumn{6}{c}{\texttt{\textbf{60s}}} \\
\cmidrule(lr){2-7}
\cmidrule(lr){8-13}
\textbf{Method}
& \texttt{\textbf{Subj.}}$\uparrow$
& \texttt{\textbf{Dyn.}}$\uparrow$
& \texttt{\textbf{Aesth.}}$\uparrow$
& \texttt{\textbf{Quality}}$\uparrow$
& \texttt{\textbf{$\Delta$IQ}}$\downarrow$
& \texttt{\textbf{VLM}}$\uparrow$
& \texttt{\textbf{Subj.}}$\uparrow$
& \texttt{\textbf{Dyn.}}$\uparrow$
& \texttt{\textbf{Aesth.}}$\uparrow$
& \texttt{\textbf{Quality}}$\uparrow$
& \texttt{\textbf{$\Delta$IQ}}$\downarrow$
& \texttt{\textbf{VLM}}$\uparrow$ \\
\midrule

\rowcolor{gray!12}
\multicolumn{13}{l}{
\textbf{\textit{Short-video supervision}}} \\

Self Forcing{\tiny~\citep{huang2025self}}
& $97.43$ & $40.42$ & $60.66$
& \cellgr $82.34$ & $5.82$ & $3.13$
& $97.27$ & $44.21$ & $56.58$
& \cellgr $81.89$ & $6.65$ & 2.69 \\

Causal Forcing{\tiny~\citep{zhu2026causal}}
& $96.10$ & $69.38$ & $57.41$
& \cellgr $82.19$ & $7.34$ & $2.45$
& $95.96$ & $59.86$ & $51.83$
& \cellgr $80.93$ & $12.97$ & 2.07 \\

LongLive{\tiny~\citep{yang2026longlive}}
& \cellte $98.03$ & $41.25$ & \cellte $63.21$
& \cellgr $83.25$ & $2.96$ & $3.90$
& \cellte $98.40$ & $37.04$ & $62.23$
& \cellgr $82.98$ & \cellte $2.69$ & 3.83 \\

Reward Forcing{\tiny~\citep{lu2026reward}}
& $97.26$ & $66.98$ & $62.42$
& \cellgr $84.36$ & \cellbg $2.58$ & $3.79$
& $97.29$ & $80.23$ & $62.08$
& \cellgr $85.54$ & \cellbg $2.45$ & 3.75 \\

MemRoPE{\tiny~\citep{kim2026memrope}}
& $97.28$ & $58.65$ & $62.33$
& \cellgr $83.95$ & $2.95$ & $3.83$
& $97.48$ & $56.20$ & $59.34$
& \cellgr $83.92$ & $3.69$ & \cellbg 3.91 \\

Rolling Forcing$^\dagger$
{\tiny~\citep{liu2026rolling}}
& $96.04$ & \cellte $93.44$ & $58.61$
& \cellgr $84.67$ & $5.58$ & $3.77$
& $95.81$ & \cellte $95.69$ & $56.75$
& \cellgr $84.78$ & $6.69$ & $3.80$ \\

Self Gradient Forcing{\tiny~\citep{zhuang2026self}}
& \cellbg $98.19$ & $51.30$ & \cellbg $63.90$
& \cellgr $84.25$ & \cellte $2.75$ & \cellbg $3.94$
& \cellbg $98.42$ & $53.06$ & \cellbg $64.09$
& \cellgr $84.15$ & $2.70$ & \cellte $3.87$ \\

\midrule
\rowcolor{gray!12}
\multicolumn{13}{l}{
\textbf{\textit{Long-video supervision}}} \\

Context Forcing{\tiny~\citep{chen2026context}}
& $97.72$ & $57.71$ & $63.05$
& \cellgr $84.16$ & $3.31$ & $3.55$
& $98.17$ & $52.69$ & $62.10$
& \cellgr $83.85$ & $3.65$ & $3.26$ \\

\rowcolor{gray!6}
\textbf{LongTake}{\scriptsize~\textbf{(Ours)}}
& $96.83$ & $90.73$ & $62.51$
& \cellte $85.79$ & $3.13$ & \cellbg $3.94$ 
& $96.78$ & $94.72$ & \cellte $62.35$ 
& \cellbg 85.96 & $3.55$ & \cellte $3.87$ \\

\rowcolor{gray!6}
\hspace{4mm}+ Hybrid DMD\gray{{\tiny~$(\lambda=0.2)$}}
& $96.32$ & \cellbg $96.30$ & $62.17$
& \cellbg $85.97$ & $3.22$ & \cellte $3.92$
& $96.43$ & \cellbg $97.36$ & $61.48$
& \cellte $85.71$ & $3.60$ & $3.86$ \\

\bottomrule
\end{tabular}%
}
\vspace{-4mm}
\end{table*}

\Needspace{17\baselineskip}

\paragraph{Long-horizon visual stability}
We assess visual stability using
\texttt{\textbf{$\Delta$IQ}}~\citep{liu2026rolling}
and \texttt{\textbf{VLM}}~\citep{kim2026memrope}.
They measure the absolute change in imaging quality
between the first and last five seconds
and exposure stability throughout the video, respectively.
We interpret \texttt{\textbf{$\Delta$IQ}} alongside imaging quality
because stable endpoint scores alone do not establish high visual quality.
LongTake matches the highest \texttt{\textbf{VLM}} score at 30s
and achieves comparable \texttt{\textbf{$\Delta$IQ}}
to Context Forcing at both durations
while producing substantially stronger dynamics.
Both LongTake variants obtain lower \texttt{\textbf{$\Delta$IQ}}
than Rolling Forcing at 30s and 60s,
although Hybrid DMD slightly increases endpoint quality changes
relative to joint DMD alone.
It also exceeds Rolling Forcing in \texttt{\textbf{VLM}}
while producing stronger dynamics.
These results support sustained dynamics with competitive visual stability
beyond the teacher supervision horizon.

\begin{remarks}
\textit{Videos with low dynamic degree show little subject or camera motion over time,
as in Figure~\ref{fig:qualitative_30s_60s}.
Such videos no longer depict the action described in the prompt
and extend duration without adding content, which undermines long-horizon generation.
Yet they can still score well, since each frame stays sharp
and consistency metrics favor static videos~\citep{liao2024devil,huang2024vbench},
so long videos need high dynamic degree as much as visual quality.
}
\end{remarks}

\paragraph{Ablation Study}

We compare generation
from shared video prefixes before DMD in Appendix~\ref{appendix:teacher-generation}.
Compared to standard TF, Long-Horizon TF improves quality at each prefix length,
while boundary metrics suggest better appearance preservation
at the immediate transition.

\begin{wraptable}[8]{r}{0.27\textwidth}
\vspace{-5mm}
\centering
\caption{
\textbf{Hybrid DMD with shared initialization.}
}
\label{tab:lambda_ablation}
\resizebox{\linewidth}{!}{%
\renewcommand{\arraystretch}{1.15}
\setlength{\tabcolsep}{5pt}
\begin{tabular}{@{}l|  ccc c@{}}
\toprule
$\lambda$
& \texttt{\textbf{Subj.}}$\uparrow$
& \texttt{\textbf{Dyn.}}$\uparrow$
& \texttt{\textbf{Aesth.}}$\uparrow$
& \texttt{\textbf{Quality}}$\uparrow$ \\
\midrule
$0.0$
& \cellbg $96.83$ & $90.73$
& $62.51$ & \cellgr  $85.79$ \\
$0.1$
& $96.78$ & $90.00$
& \cellbg $62.64$ & \cellgr  $85.83$ \\
$0.2$
& $96.32$ & \cellbg $96.30$
& $62.17$ & \cellbg $85.97$ \\
$0.3$
& $96.38$ & $95.57$
& $61.49$ & \cellgr  $85.85$ \\
$0.4$
& $96.39$ & $95.42$
& $61.91$ & \cellgr  $85.61$ \\
\bottomrule
\end{tabular}%
}
\end{wraptable}

\Cref{tab:lambda_ablation} evaluates supervision of later frames
on 30s generation.
All settings use the same Long-Horizon TF initialization
and 1,200 distillation iterations.
Joint DMD retains 5s supervision throughout,
while Hybrid DMD extends the rollout after 800 iterations.
With $\lambda=0.2$, Hybrid DMD achieves the highest dynamic degree
and Quality score among the tested settings,
with modest reductions in subject consistency and aesthetic quality.
The 60s results in~\Cref{tab:main_results} show stronger dynamics
with a small decrease in Quality relative to joint DMD,
making Hybrid DMD an extension with a motion--quality tradeoff. Appendix~\ref{appendix:qualitative-results}
illustrate the additional scene progression with coherent subject appearance.

\section{Conclusion}

We introduced~\ours, a two-stage training pipeline
for long-horizon AR video generation.
Long-Horizon TF learns from curated real long videos
by predicting later frames conditioned on long ground-truth video prefixes.
The resulting model directly initializes self-rollout DMD
without a separate few-step initialization stage.
Under the standard DMD, this initialization yields higher dynamic degree
than short TF initialization at comparable aesthetic quality.
Hybrid DMD further reuses the model as a teacher
for later frames of the student self-rollout,
achieving the highest dynamic degree among evaluated methods.
Both variants lie on the Pareto front of dynamic degree and aesthetic quality
on 30s and 60s rollouts. These results demonstrate the value of long-video supervision
for learning to sustain dynamics over long-horizon.

\paragraph{Limitations}
The effectiveness of Hybrid DMD depends on the predictive quality
of the Long-Horizon AR teacher used to supervise later frames.
Future work could incorporate error-recycling fine-tuning,
as in SVI~\citep{li2026stable},
to improve teacher robustness to errors accumulated
in the video prefix during student self-rollout.
Evaluation is limited to a $\ttb{1.3B}$ backbone and rollouts up to 60s,
leaving performance with larger backbones and longer rollouts untested.

\newpage
\clearpage

\subsection*{AI use statement}

The research ideas and proposed methods originated with the authors.
We used generative AI tools to help refine author-developed hypotheses,
assist with implementing the proposed methods and experimental procedures,
support dataset cleaning and reformatting,
and assist with interpreting experimental results.
Additionally, we used these tools for manuscript drafting and revision,
as well as literature search, discovery, and summarization.
We also used Gemini 3.1 Pro Preview to evaluate exposure stability
in generated videos, following the protocol described
in Appendix~\ref{appendix:evaluation-protocol}.
The authors reviewed and revised the AI-assisted material
and take full responsibility for the final content of this work,
including its methodology, implementation, claims,
and reported results.

\subsection*{Reproducibility statement}

Appendix~\ref{appendix:method details} documents
student rollout, gradient replay, and Hybrid DMD updates.
Appendix~\ref{appendix:training-configuration} specifies
initialization, optimization, cache configurations, and sampling.
Appendix~\ref{appendix:training-data} describes the training corpus and curation,
Appendix~\ref{appendix:baseline-configuration} lists
baseline checkpoints and generation settings,
and Appendix~\ref{appendix:evaluation-protocol} defines
the evaluation prompts, metrics, and score aggregation.

\bibliography{main}
\bibliographystyle{iclr2027_conference}

\newpage
\clearpage

\appendix
\section{Extended Related Work}
\label{appendix:Related work}

\paragraph{Autoregressive Video Diffusion.}
Autoregressive video diffusion generates videos
by conditioning each chunk on a preceding video prefix.
Diffusion Forcing~\citep{chen2024diffusion}
combines next-token prediction with sequence diffusion
through independent noise levels,
while CausVid~\citep{yin2025slow}
distills a bidirectional teacher into a few-step causal generator.
Self Forcing~\citep{huang2025self}
addresses the training--inference gap
by training under student self-rollout.
Causal Forcing~\citep{zhu2026causal}
and Causal Forcing++~\citep{zhao2026causal}
improve student initialization through causal ODE distillation
and causal consistency distillation, respectively.
Self Gradient Forcing~\citep{zhuang2026self}
uses gradient replay to train memory writing for subsequent generation
and evaluates direct initialization from a TF-trained model.
Building on direct TF initialization,
LongTake studies how supervision from real long videos
strengthens the AR model before self-rollout DMD.

\paragraph{Long-Horizon Video Generation.}
Existing methods extend video generation
through sampling and context management.
FIFO-Diffusion~\citep{kim2024fifo} uses diagonal denoising,
while StreamingT2V~\citep{henschel2025streamingt2v}
introduces short- and long-term memory.
FramePack~\citep{zhang2025frame}
and PackForcing~\citep{mao2026packforcing}
compress preceding video content,
and Deep Forcing~\citep{yi2026deep}
and Infinity-RoPE~\citep{yesiltepe2026infinity}
adapt cache management and temporal positions during inference.
LongLive~\citep{yang2026longlive},
Rolling Forcing~\citep{liu2026rolling},
and MemRoPE~\citep{kim2026memrope}
further improve extended AR rollouts.
AdaState~\citep{dalva2026adastate}
addresses motion stagnation
by replacing a fixed anchor with an evolving state.
LongTake focuses on learning to predict later frames
under conditioning KV-cache states constructed
from real long-video prefixes. Such training could complement this line of work
by helping the AR model better use the information
retained in the KV-cache during long-horizon generation.

\paragraph{Long-Horizon Supervision and Distillation.}
Long-video data provides supervision
for scene evolution beyond short clips.
LVD-2M~\citep{xiong2024lvd}
and Presto~\citep{yan2025long}
emphasize dynamic long-take videos,
while Long Context Tuning~\citep{guo2025long}
learns dependencies across extended scenes.
Resampling Forcing~\citep{guo2026end}
combines longer-video training
with robustness to errors in generated video prefixes.
For distillation,
Self-Forcing++~\citep{cui2026selfforcing}
applies DMD to short windows sampled from extended student rollouts
using a short-video teacher.
MMM~\citep{cai2026mode}
combines supervised flow matching on long videos
with sliding-window distribution matching
to a short-video teacher through a shared encoder.
Context Forcing~\citep{chen2026context}
trains a long-context teacher on real long videos
and applies prefix-conditioned distribution matching.
Context-Matched Distillation~\citep{bandyopadhyay2026context}
aligns causal teacher scoring with the student self-rollout prefix
and uses the teacher to initialize the student.
LongTake examines how real long-video supervision
improves direct TF initialization under short-horizon joint DMD.
The resulting AR teacher can then supervise later frames
through Hybrid DMD under student self-rollout.

\subsection{Comparison with closely related work.}
\label{sec:Comparison with existing work.}

\paragraph{Long-Horizon TF initialization.}
Causal Forcing~\citep{zhu2026causal}
and Causal Forcing++~\citep{zhao2026causal}
introduce separate causal ODE and consistency distillation stages
before self-rollout DMD.
Direct initialization from a TF-trained model
is also evaluated in these studies
and in SGF~\citep{zhuang2026self}.
To strengthen this direct initialization,
LongTake uses supervision from real long videos.
We curate a long-video dataset
and fine-tune the AR model through Long-Horizon TF,
pairing KV-cache states from long video prefixes
with subsequent ground-truth target frames.
In~\Cref{fig:video_difference_with_scores},
we compare short-horizon TF, short-horizon TF followed by causal CD,
and Long-Horizon TF under the same subsequent 5s joint DMD configuration.
Long-Horizon TF on curated real long videos yields stronger dynamics than short-horizon TF with comparable visual quality, while retaining higher aesthetic quality than causal CD initialization.
The resulting two-stage pipeline achieves strong long-horizon generation
without a separate few-step initialization stage.
These gains show the value of long-video supervision before extending the DMD horizon.

\paragraph{Context Forcing.}
Context Forcing~\citep{chen2026context}
also curates real long videos to train a long-context AR teacher
and uses this teacher for contextual DMD over extended student rollouts.
LongTake examines the benefit of this supervision
as a direct student initialization while keeping subsequent joint DMD at 5s.
The resulting gains show that long-video supervision
can improve long-horizon generation
before the AR teacher scores extended student rollouts.
When distillation is extended through Hybrid DMD,
the two methods also differ in their target distributions.
Context Forcing matches a target window conditioned on a shared video prefix,
whereas our conditional DMD matches the next chunk
under the student self-rollout prefix updated before each chunk.

\paragraph{Context-Matched Distillation.}
Concurrent work, Context-Matched Distillation~\citep{bandyopadhyay2026context},
trains a multi-step causal teacher with Diffusion Forcing
and directly initializes the student from its weights,
without a separate ODE or consistency distillation stage.
Its Prefix Scoring conditions teacher scores
on a conditioning KV-cache constructed from student-generated video prefixes,
while Prefix Corruption stabilizes supervision under imperfect prefixes.
LongTake targets the teacher gap with Long-Horizon TF on curated real long videos and shows stronger direct initialization under the same 5s joint DMD procedure.
These gains arise before changing the bidirectional scoring
or extending the DMD horizon.
Hybrid DMD adds AR supervision of later frames
while retaining bidirectional joint supervision of the initial window.

\section{Implementation Details}
\label{appendix:method details_0}

\subsection{Parallel Cache Computation}
\label{appendix:parallel cache prefill}

\begin{figure}[t]
\centering
\begin{minipage}[t]{1.\textwidth}
\vspace{0pt}
\centering
    \includegraphics[width=1.\linewidth]{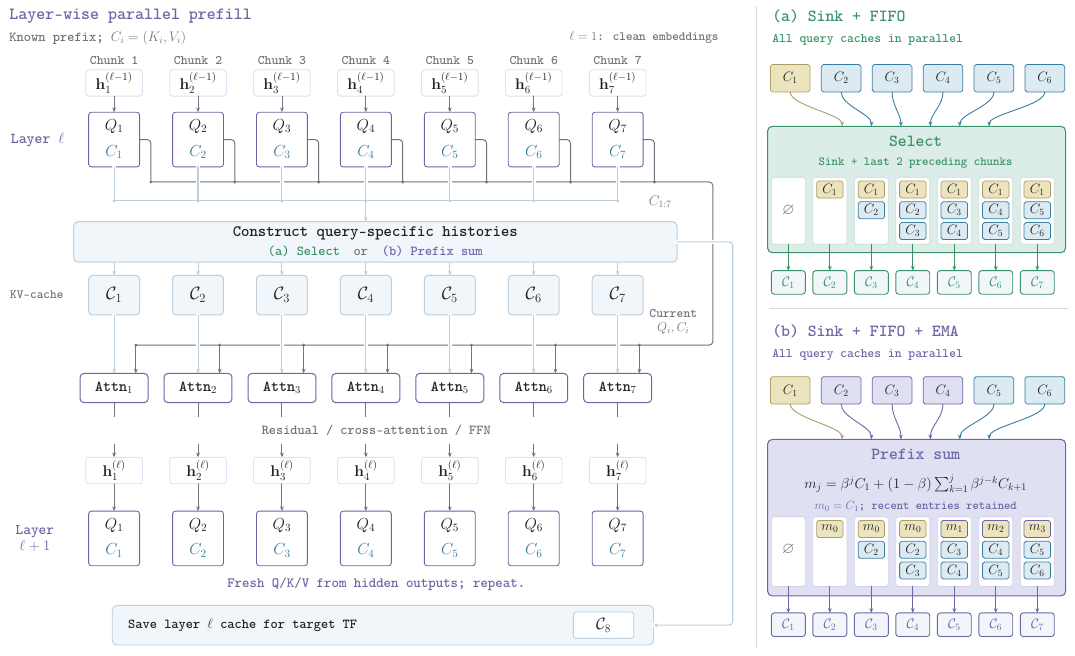}
\caption{
\textbf{Parallel cache construction for Long-Horizon TF.}
(Left) At each transformer layer, all known prefix chunks
are projected together to $Q_i$ and $C_i=(K_i,V_i)$.
For each query, the cache operator constructs
a conditioning KV-cache $\mathcal C_i$,
which enters causal attention together with
the current chunk's Q/K/V.
The resulting hidden states feed the next layer.
(Right) Two alternative operators construct the KV-cache states
of all queries in parallel.
Operator (a) selects the initial sink and the most recent preceding chunks,
and operator (b) additionally merges evicted entries into the sink
through an associative weighted prefix scan.
The example uses seven chunks with one sink and two recent slots.
After prefill, $\mathcal C_8$ at each layer is saved for the
subsequent target-window TF pass.
Parallelism is across chunks within each layer,
while transformer layers remain sequential.
}
\label{fig:parallel-prefill}
\end{minipage}
\end{figure}
 
Long-Horizon TF constructs a conditioning KV-cache from a clean
ground-truth video prefix before evaluating the target-window loss.
Since all prefix chunks are known, cache construction can process
all chunks in parallel within each layer.
\Cref{fig:parallel-prefill} illustrates this computation for two
cache-construction rules: sink retention with FIFO eviction, used in
our Long-Horizon TF implementation, and a fixed-coefficient EMA extension.
 
\paragraph{Shared layer-wise computation.}
Let the video prefix contain $P$ chunks, and let
$\mathbf h_i^{(\ell-1)}$ denote the hidden states of chunk $i$
entering transformer layer $\ell$.
All chunks are projected together to obtain
\begin{equation}
    (Q_i^{(\ell)},K_i^{(\ell)},V_i^{(\ell)})
    = \operatorname{Proj}_{\ell}(\mathbf h_i^{(\ell-1)}),
    \qquad i=1,\ldots,P,
\end{equation}
where $\operatorname{Proj}_{\ell}$ includes the corresponding
normalization and projection operations.
We write $C_i^{(\ell)}=(K_i^{(\ell)},V_i^{(\ell)})$ for one
chunk's projected K/V pair, and $\mathcal C_i^{(\ell)}$ for the
conditioning KV-cache available \emph{before} processing chunk $i$.
The cache operator constructs each $\mathcal C_i^{(\ell)}$
from $C_1^{(\ell)},\ldots,C_{i-1}^{(\ell)}$.
Each query then attends to its conditioning KV-cache
and its own current K/V,
\begin{equation}
    A_i^{(\ell)}
    = \operatorname{Attn}\!\left(
        Q_i^{(\ell)};\,
        \mathcal C_i^{(\ell)}\Vert C_i^{(\ell)}
      \right),
\end{equation}
where $\Vert$ concatenates keys and values along the token dimension.
Constructing these KV-cache states requires no attention output
from another chunk at the same layer,
so all query groups can be evaluated in parallel
while preserving chunk causality.
The remaining transformer operations produce $\mathbf h_i^{(\ell)}$,
from which layer $\ell+1$ computes fresh projections
after layer $\ell$ completes.
 
\paragraph{Sink + FIFO via parallel selection.}
For clarity, consider one sink chunk and a capacity of $r$ recent
chunks, with $r=2$ in \Cref{fig:parallel-prefill}.
Long-Horizon TF uses $r=5$ with three-frame chunks,
which retains three sink frames and 15 recent frames.
We omit layer superscripts below.
For $i\geq2$, the recent indices are
\begin{equation}
    \mathcal R_i
    = \{j:\max(2,i-r)\leq j\leq i-1\},
    \qquad
    \mathcal C_i^{\mathrm{FIFO}}
    = [C_1]\Vert[C_j]_{j\in\mathcal R_i},
\end{equation}
with $\mathcal C_1^{\mathrm{FIFO}}=\varnothing$.
An empty index interval contributes no entries.
The retained indices depend only on the query position and cache capacity,
so the KV-cache states of all queries are obtained
by parallel indexed selection from the shared K/V bank,
without sequentially updating a cache for each query.
For example, query $7$ uses $[C_1,C_5,C_6]$ when $r=2$,
since $C_2$ to $C_4$ have been evicted.
 
\paragraph{Sink + FIFO + EMA via parallel prefix scan.}
The EMA extension additionally merges evicted entries into the sink
with a fixed coefficient $\beta\in[0,1)$~\citep{lu2026reward}.
With equal-sized chunks,
\begin{equation}
    m_0=C_1,\qquad
    m_j=\beta m_{j-1}+(1-\beta)C_{j+1}
       =\beta^jC_1+(1-\beta)
         \textstyle{\sum_{k=1}^{j}}\beta^{j-k}C_{k+1}.
\end{equation}
The same weights are applied separately to keys and values.
Before query $i\geq2$, exactly $e_i=\max(0,i-r-2)$ non-sink
chunks have been evicted, giving
\begin{equation}
    \mathcal C_i^{\mathrm{EMA}}
    = [m_{e_i}]\Vert[C_j]_{j\in\mathcal R_i},
    \qquad \mathcal C_1^{\mathrm{EMA}}=\varnothing.
\end{equation}
Although the recurrence is written sequentially, its updates are
affine maps $m\mapsto am+b$ whose ordered composition is associative,
\begin{equation}
    (a_2,b_2)\circ(a_1,b_1)
    = (a_2a_1,\,a_2b_1+b_2).
\end{equation}
An associative prefix scan~\citep{blelloch1990prefix} therefore computes
all required sink states in parallel with logarithmic scan depth.
Each query selects its corresponding sink state and recent entries.
For example, query $7$ uses $[m_3,C_5,C_6]$ when $r=2$,
where $m_3$ merges $C_2$ to $C_4$ into the sink.
 
\paragraph{Saved cache and target prediction.}
At each layer, the same operator constructs $\mathcal C_{P+1}^{(\ell)}$,
the cache after all $P$ prefix chunks.
For the seven-chunk example, this is $[C_1,C_6,C_7]$ for FIFO
or $[m_4,C_6,C_7]$ for EMA.
Collecting these final caches across layers yields the conditioning
KV-cache $\ttb{c}_{\ttb{gt}}^s$ used in~\Cref{alg:long-teacher}.
For matching query-specific cache selection, aggregation, and positional
treatment, the layer-wise schedule is mathematically equivalent to
sequential prefix processing; numerical differences can arise from
floating-point accumulation order.
The prefix pass runs without gradient tracking, and the saved cache
is detached.
A second, two-stream forward pass evaluates TF losses over the fixed
target window: clean chunks provide preceding ground-truth context,
while noisy chunks predict target velocities.
Gradients remain enabled through the clean-context computation within
this target window.

\paragraph{Bounded context within the target window.}
The saved prefix cache supplies the initial history for target prediction.
For each target chunk, the same sink-retention and FIFO rule is applied
to the saved history and preceding clean target chunks in temporal order.
The original sink is retained, while the recent slots contain at most
the five most recent non-sink chunks.
Thus, each query uses at most 18 past latent frames plus its current
three-frame chunk, giving a maximum attention window of 21 latent frames.
The clean and noisy query streams use the same past-history selection,
but each supplies its own current K/V.
In particular, a noisy query cannot access the clean K/V of its own
target chunk or any future chunk.
These query-specific selections implement the cache updates within the
packed two-stream forward pass.

\paragraph{Bounded positional encoding.}
In our FIFO implementation, keys are stored before the rotary positional
transformation and selected separately for each query.
If a query retains $h\leq18$ past latent frames, the concatenated history
and current keys receive temporal RoPE positions $0,\ldots,h+2$,
while the current queries receive positions $h,h+1,h+2$.
At full capacity, the sink occupies positions $0$--$2$, the recent frames
occupy $3$--$17$, and the current chunk occupies $18$--$20$.
This reindexing keeps the temporal positions within the 21-frame range
regardless of the target's original position in the long video.
For $s=0$, no prefix pass is needed and the computation reduces
to standard TF.

\subsection{Hybrid DMD}
\label{appendix:method details}

We implement the joint objective
in~\eqref{eq:joint-dmd}
using Self Gradient Forcing~\citep{zhuang2026self}.
We describe how gradient replay computes the student updates
and how these updates extend to Hybrid DMD.
Model configurations and training settings
are provided in Appendix~\ref{appendix:training-configuration}.

\paragraph{Student rollout and gradient replay.}
We first perform student self-rollout
without gradient tracking.
On each device,
we uniformly select one denoising step
and share this selection across all chunks.
Although the rollout completes all denoising steps,
we record the noisy inputs only at the selected step,
together with the corresponding inputs for encoding the video prefix.

We then replay these recorded inputs
through a packed two-stream student forward pass.
The prefix stream encodes
the recorded student self-rollout prefix,
while the noisy stream predicts each current chunk.
Causal masking prevents the noisy stream
from accessing current or future target frames
in the prefix stream.
Let $\hat{\bx}_{\theta}^{<\green{N}}$
denote the resulting clean predictions
over the initial joint window.

Although both recorded input streams are detached,
replay recomputes the video-prefix representations,
allowing gradients to reach
the student parameters that encode the prefix.
The update therefore differentiates through replay
without backpropagating through the preceding student self-rollout.

\begin{figure}[t]
\centering
\begin{minipage}[t]{1.\textwidth}
\vspace{0pt}  
\centering
    \includegraphics[width=0.8\linewidth]{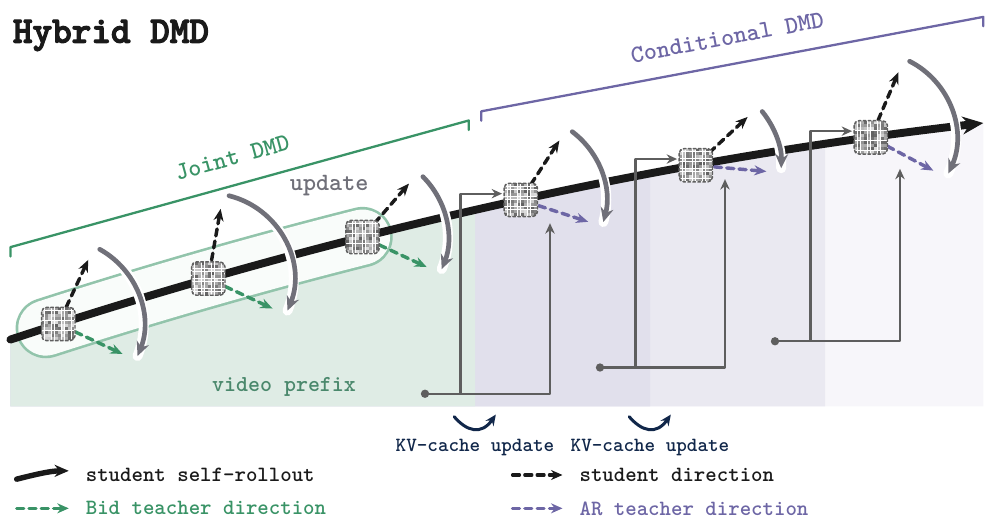}
\caption{\textbf{Hybrid DMD.} A bidirectional teacher jointly supervises
the initial window. The Long-Horizon AR teacher supervises later frames
using the conditioning KV-cache updated during student self-rollout
for each target chunk.}
\label{fig:hybrid_dmd}
\end{minipage}
\end{figure}

\paragraph{Hybrid DMD update.}
We write the two contributions
to the objective in~\eqref{eq:hybrid-dmd} as
$\calJ_{\ttb{DMD}}^{\ttb{Bi}}$
and $\calJ_{\ttb{DMD}}^{\ttb{AR}}$.
The student update combines these contributions
over the initial window and later frames as
\begin{equation}
    \widehat{\nabla}_{\theta}
    \calJ_{\ttb{Hybrid}}
    =
    \widehat{\nabla}_{\theta}
    \calJ_{\ttb{DMD}}^{\ttb{Bi}}
    +
    \lambda
    \widehat{\nabla}_{\theta}
    \calJ_{\ttb{DMD}}^{\ttb{AR}}.
    \label{eq:hybrid-dmd-gradient}
\end{equation}
Here, $\widehat{\nabla}_{\theta}$
denotes the update estimated
using detached self-rollout inputs
and SGF replay.
The following expressions describe the joint and conditional contributions
in terms of score differences.

\paragraph{Joint contribution.}
Let $s_{\psi}^{\ttb{Bi}}$
denote the trainable joint fake-score model
and $s^{\ttb{Bi}}$
the frozen bidirectional teacher score.
Both models evaluate
the complete initial noisy window
$\bx_t^{<\green{N}}$
at the same timestep.
The joint contribution follows from their score difference over this window
\[
    \widehat{\nabla}_{\theta}
    \calJ_{\ttb{DMD}}^{\ttb{Bi}}
    =
    \bbE
    \Bigg[
        w(t)
        \Big( s_{\psi}^{\ttb{Bi}}
            (t,\bx_t^{<\green{N}}) -
            s^{\ttb{Bi}}
            (t,\bx_t^{<\green{N}})
        \Big)^{\top}
        \frac{
            \partial\bx_t^{<\green{N}}
        }{
            \partial\theta
        }
    \Bigg].
\label{eq:bi-dmd-gradient}
\]
The expectation is over
student self-rollout,
the sampled denoising step for replay,
and the score-evaluation timestep and noise.
Although generation is autoregressive,
both score models evaluate the initial window jointly
rather than scoring each target chunk under a separately updated prefix.

\paragraph{Conditional contribution.}
For each target chunk beyond the initial window,
we evaluate the conditional fake-score model
$s_{\psi}^{\ttb{AR}}$
and the frozen Long-Horizon AR teacher
$s_{\phi^\star}^{\ttb{AR}}$
under the same student self-rollout prefix.
Using half-open temporal intervals,
the corresponding chunk indices satisfy
$\green{N}\leq i<\purple{M}$.
Averaging over these target chunks gives the conditional contribution
\[
    \widehat{\nabla}_{\theta}
    \calJ_{\ttb{DMD}}^{\ttb{AR}}
    ={}&
    \frac{1}{\purple{M}-\green{N}}
    \sum_{i=\green{N}}^{\purple{M}-1}
    \bbE
    \Bigg[
        w(t_i)
        \Big(
            s_{\psi}^{\ttb{AR}}
            \left(
                t_i,\bx_{t_i}^{i},
                \ttb{c}_{\ttb{self}}^{i}
            \right)
            -
            s_{\phi^\star}^{\ttb{AR}}
            \left(
                t_i,\bx_{t_i}^{i},
                \ttb{c}_{\ttb{self}}^{i}
            \right)
        \Big)^{\top}
        \frac{
            \partial\bx_{t_i}^{i}
        }{
            \partial\theta
        }
    \Bigg].
\label{eq:ar-dmd-gradient}
\]
Each expectation additionally includes
the sampled student self-rollout prefix.
The cache notation identifies
the shared conditioning video prefix,
which each score model encodes
using its own parameters.
Noise is applied only to the target chunk,
so both conditional scores use an unperturbed student self-rollout prefix
while evaluating the same noisy target chunk.

We treat the sampled student self-rollout prefix
as fixed when computing this update.
Thus, gradients do not propagate through
the generation of preceding chunks,
but student replay retains gradients
through the video-prefix representations
recomputed from this fixed prefix.
Averaging over target chunks beyond the initial window
then gives the conditional update used during training.

\section{Experimental Details}
\label{appendix:experimental details}

\subsection{Training Configuration}
\label{appendix:training-configuration}

\paragraph{Model initialization.}
We initialize Long-Horizon TF from the released chunk-wise AR diffusion checkpoint of Causal Forcing\footnote{\url{https://huggingface.co/zhuhz22/Causal-Forcing/blob/main/chunkwise/ar_diffusion.pt} under Apache-2.0}~\citep{zhu2026causal}, based on $\ttb{Wan2.1}$-$\ttb{T2V}$-$\ttb{1.3B}$~\citep{wan2025}.
After 3,000 additional updates, the resulting model initializes the student and serves as the frozen Long-Horizon AR teacher.
LongTake then trains the student with joint DMD over 21 latent frames for 1,200 iterations using Self Gradient Forcing~\citep{zhuang2026self}.
For Hybrid DMD, training branches after 800 joint DMD iterations and proceeds on 42-latent-frame rollouts for another 400 iterations.
In these extended rollouts, a frozen $\ttb{Wan2.1}$-$\ttb{T2V}$-$\ttb{14B}$ teacher jointly supervises the first 21 latent frames, while the Long-Horizon AR teacher conditionally supervises the remaining 21.

\paragraph{Optimization.}
Table~\ref{tab:appendix-training} summarizes the AdamW~\citep{loshchilov2018decoupled}
configuration for both training stages.
During distillation, we update the fake-score models every iteration and the student every five iterations.
We initialize parameter EMA after iteration 200 and update it after each subsequent student update with decay 0.99.
For each variant, we use the same final EMA student and KV-cache size for both 30s and 60s evaluation, keeping model parameters and cache capacity fixed.

\paragraph{Initialization comparison.}
\label{appendix:initialization-comparison}
In~\Cref{fig:video_difference_with_scores},
Short TF, Short TF + CD, and Long TF use the same 5s joint DMD configuration in Table~\ref{tab:appendix-training}, differing only in student initialization.
Short TF and Short TF + CD start from the short-horizon AR teacher and the causal CD checkpoint released by Causal Forcing~\citep{zhu2026causal, zhao2026causal}, respectively.
For Short TF, we evaluate the released SGF EMA checkpoint~\citep{zhuang2026self} trained with this configuration, while Long TF starts from our Long-Horizon TF model trained on the curated long-video dataset.
We evaluate all three models on 30s rollouts using the same protocol, so the comparison measures the effect of initialization under a shared distillation procedure.

\paragraph{Long-Horizon TF.}
Each update supervises a target window of 21 latent frames following a ground-truth video prefix.
We sample the prefix length $s$ in multiples of three according to Table~\ref{tab:appendix-prefix}.
After selecting a bin and sampling a valid prefix length uniformly within that bin, we sample a compatible video length in proportion to its number of clips and select eight distinct clips for the eight devices.
Although clips are distinct within each microbatch, sampling across microbatches uses replacement.
Each clip contains at least $s+21$ latent frames, and the maximum prefix length is 99, so supervision reaches 120 latent frames with the loss window fixed at 21.

For $s>0$, we construct the conditioning KV-cache with one parallel pass through the transformer without gradient tracking.
A second pass then jointly processes the clean context and noisy target streams.
The conditioning KV-cache is detached, while gradients are retained through the clean-context computation within the target window.
For $s=0$, no prefix cache is needed, so standard TF processes the clean context and noisy target together in a single transformer pass.

\begin{table}[t]
\centering
\caption{\textbf{Training configuration.}
LongTake uses 1,200 joint DMD iterations.
The Hybrid DMD variant uses 800 joint DMD iterations
followed by 400 Hybrid DMD iterations.}
\label{tab:appendix-training}
\small
\setlength{\tabcolsep}{4pt}
\begin{tabular}{lccc}
\toprule
Setting & Long-Horizon TF & Joint DMD & Hybrid DMD \\
\midrule
Iterations & 3,000 & 1,200 / 800 & 400 \\
Generator learning rate & $5\times10^{-7}$ & $2\times10^{-6}$ & $2\times10^{-6}$ \\
Fake-score learning rate & --- & $4\times10^{-7}$ & $4\times10^{-7}$ \\
Effective batch size & 64 & 8 & 8 \\
Gradient accumulation & 8 & 1 & 1 \\
Optimizer & AdamW & AdamW & AdamW \\
$(\beta_1,\beta_2)$ & $(0,0.999)$ & $(0,0.999)$ & $(0,0.999)$ \\
Weight decay & 0.01 & 0.01 & 0.01 \\
Learning-rate schedule & Constant & Constant & Constant \\
Gradient clipping & 10 & 10 & 10 \\
Parameter EMA decay & --- & 0.99 & 0.99 \\
\bottomrule
\end{tabular}
\end{table}

\begin{table}[t]
\centering
\caption{Prefix sampling for Long-Horizon TF. Prefix lengths are measured in latent frames and sampled in multiples of three. Probabilities refer to the complete training mixture.}
\label{tab:appendix-prefix}
\small
\begin{tabular}{lcc}
\toprule
Source & Prefix length $s$ & Probability \\
\midrule
Causal Forcing clean GT & 0 & 0.10 \\
OpenVidHD & 0--15 & 0.20 \\
OpenVidHD & 18--39 & 0.45 \\
OpenVidHD & 42--60 & 0.20 \\
OpenVidHD & 63--99 & 0.05 \\
\bottomrule
\end{tabular}
\end{table}

\paragraph{Cache configuration.}
Long-Horizon TF, the AR teacher, and the conditional fake-score model retain three sink frames and 15 recent frames.
Together with the current three-frame chunk, these form an attention window of 21 latent frames.
Following~\citep{zhuang2026self}, the student uses three sink frames and six recent frames, giving a 12-frame attention window including the current chunk.

\paragraph{Distillation and sampling.} 
The four-step student uses denoising timesteps
$t\in\{1,0.9375,0.8333,0.625\}$.
Using the guidance convention $u+\omega(c-u)$, the bidirectional and AR teachers use $\omega=4$ and $\omega=3$, respectively.
For Hybrid DMD, the generator loss is the mean joint surrogate plus $\lambda=0.2$ times the mean conditional surrogate over the seven target chunks covering the later frames.
Both fake-score models use flow regression on fresh student samples, with equal weights for their two losses.
For the positive-$\lambda$ comparisons, runs branch from the same state after 800 joint-DMD iterations and receive 400 Hybrid-DMD iterations.
The $\lambda=0$ control uses joint DMD for all 1,200 iterations, keeping the total distillation budget fixed across Table~\ref{tab:lambda_ablation}.

\Needspace{13\baselineskip}
\subsection{Training Data}
\label{appendix:training-data}

\begin{wrapfigure}[10]{r}
{0.41\linewidth}
\begin{minipage}[t]{0.4\textwidth}
\vspace{-7mm}
\centering
    \includegraphics[width=1.\linewidth]{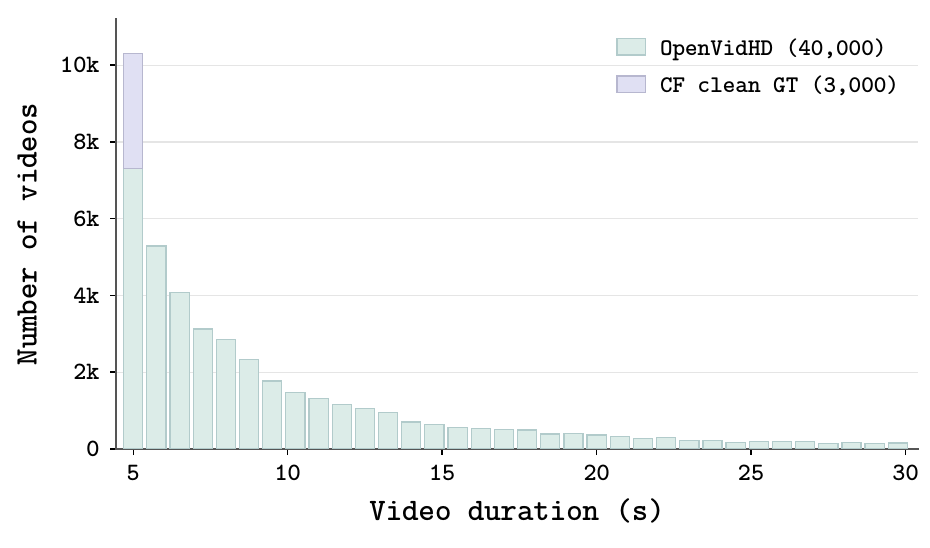}
\vspace{-5mm}
\caption{\textbf{Dataset duration.}}
\label{fig:dataset_duration}
\end{minipage}
\end{wrapfigure}

\paragraph{Video corpus.}
Long-Horizon TF uses 43,000 video--text pairs comprising 40,000 OpenVidHD clips\footnote{\url{https://huggingface.co/datasets/nkp37/OpenVid-1M} under CC BY 4.0}~\citep{nan2025openvidm} and 3,000 short clips from the released Causal Forcing clean-synthesized data\footnote{\url{https://huggingface.co/zhuhz22/Causal-Forcing-data}}~\citep{zhu2026causal}.
OpenVidHD clips contain 21--120 latent frames, corresponding to approximately 5--30 seconds at 16 FPS.
The short subset contains 21 latent frames per clip, and we retain the original captions and clean-GT prompts for both sources.
During training, we sample OpenVidHD with probability 0.9 and the short subset with probability 0.1.

\paragraph{OpenVidHD curation.}
We begin with 433,509 registry entries from the OpenVidHD~\citep{nan2025openvidm} subset of OpenVid-1M.
We require a source frame rate of at least 16 FPS and a candidate latent length between 21 and 123, computed from the source video metadata as
\[
\left\lfloor
\frac{16\,(\text{source frame count})/(\text{source FPS})+3}{4}
\right\rfloor.
\]
These criteria retain 243,010 candidates.
Since OpenVidHD clips generally exhibit limited dynamics, we select 40,000 videos from these candidates to match the motion statistics of the short-horizon TF reference data while filtering for visual quality and temporal consistency.

\subsection{Baseline Configuration}
\label{appendix:baseline-configuration}

\paragraph{Baselines.}
Self Forcing~\citep{huang2025self} uses the released DMD EMA checkpoint.
Causal Forcing~\citep{zhu2026causal} uses its released chunk-wise DMD generator.
LongLive~\citep{yang2026longlive} uses the released base model with its rank-256 LoRA and a 12-frame window containing three sink frames, six recent frames, and the current chunk.
Reward Forcing~\citep{lu2026reward} uses a nine-frame window containing three sink frames, three recent frames, and the current chunk.
It updates the sink with weights 0.999 on the previous sink and 0.001 on evicted content.
MemRoPE~\citep{kim2026memrope} uses 12-frame window contains three sink frames, two EMA memory frames, four recent frames, and the current chunk, with new-content weights 0.01 and 0.1 for the two memory frames.
Rolling Forcing$^\dagger$ denotes the official long-video model released by Causal Forcing~\citep{zhu2026causal}, which combines the Rolling Forcing framework~\citep{liu2026rolling} with causal ODE initialization.
We use this released four-step model with its 24-frame rolling-window configuration.
SGF~\citep{zhuang2026self} uses the released chunk-wise TF-initialized EMA checkpoint with three sink frames, six recent frames, and the current chunk.
Context Forcing~\citep{chen2026context} uses its released EMA checkpoint with native Slow-Fast Memory.
This configuration retains three sink frames within a 21-frame cache and selects keyframes from the video prefix by similarity.

\begin{table*}[t]
\centering
\caption{\textbf{Full evaluation on 30s and 60s autoregressive rollouts.}
\texttt{\textbf{Quality}} aggregates seven video-quality dimensions
using the standard VBench normalization and weights.
LongTake uses Long-Horizon TF and joint DMD on 5s rollouts. The \best{best} and \secondbest{second-best} results are highlighted.}
\resizebox{\textwidth}{!}{%
\renewcommand{\arraystretch}{1.2}
\setlength{\tabcolsep}{4pt}
\begin{tabular}{@{}lccccccc|c@{}}
\toprule
& \multicolumn{8}{c}{\texttt{\textbf{30s}}} \\
\cmidrule(lr){2-9}
\textbf{Method}
& \texttt{\textbf{Subj.}}$\uparrow$
& \texttt{\textbf{Backg.}}$\uparrow$
& \texttt{\textbf{Motion}}$\uparrow$
& \texttt{\textbf{Dyn.}}$\uparrow$
& \texttt{\textbf{Aesth.}}$\uparrow$
& \texttt{\textbf{Imaging}}$\uparrow$
& \texttt{\textbf{Flicker}}$\uparrow$
& \texttt{\textbf{Quality}}$\uparrow$ \\
\midrule

\rowcolor{gray!12}
\multicolumn{9}{l}{\textbf{\textit{Short-video supervision}}} \\

Self Forcing{\tiny~\citep{huang2025self}}
& $97.43$ & $96.50$ & $98.55$ & $40.42$ & $60.66$
& $69.27$ & $98.86$ & \cellgr $82.34$ \\

Causal Forcing{\tiny~\citep{zhu2026causal}}
& $96.10$ & $95.76$ & $96.99$ & $69.38$ & $57.41$
& $66.66$ & $98.24$ & \cellgr $82.19$ \\

LongLive{\tiny~\citep{yang2026longlive}}
& \cellte $98.03$ & \cellte $97.01$ & \cellbg $98.78$
& $41.25$ & \cellte $63.21$
& $69.50$ & \cellte $99.06$ & \cellgr $83.25$ \\

Reward Forcing{\tiny~\citep{lu2026reward}}
& $97.26$ & $96.49$ & $98.15$ & $66.98$ & $62.42$
& $69.64$ & $98.61$ & \cellgr $84.36$ \\

MemRoPE{\tiny~\citep{kim2026memrope}}
& $97.28$ & $96.50$ & $98.19$ & $58.65$ & $62.33$
& $70.62$ & $98.77$ & \cellgr $83.95$ \\

Rolling Forcing$^{\dagger}${\tiny~\citep{liu2026rolling}}
& $96.04$ & $95.44$ & $97.48$ & \cellte $93.44$ & $58.61$
& $69.08$ & $97.96$ & \cellgr $84.67$ \\

Self Gradient Forcing{\tiny~\citep{zhuang2026self}}
& \cellbg $98.19$ & \cellbg $97.02$ & \cellte $98.75$
& $51.30$ & \cellbg $63.90$
& \cellbg $71.55$ & $98.56$ & \cellgr $84.25$ \\

\midrule
\rowcolor{gray!12}
\multicolumn{9}{l}{\textbf{\textit{Long-video supervision}}} \\

Context Forcing{\tiny~\citep{chen2026context}}
& $97.72$ & $96.86$ & $98.47$ & $57.71$ & $63.05$
& $68.73$ & \cellbg $99.14$ & \cellgr $84.16$ \\

\rowcolor{gray!6}
\textbf{LongTake}{\scriptsize~\textbf{(Ours)}}
& $96.83$ & $96.11$ & $97.38$ & $90.73$ & $62.51$
& \cellte $70.68$ & $98.59$ & \cellte $85.79$ \\

\rowcolor{gray!6}
\hspace{4mm}+ Hybrid DMD\gray{{\tiny~$(\lambda=0.2)$}}
& $96.32$ & $95.62$ & $97.26$ & \cellbg $96.30$ & $62.17$
& $70.65$ & $98.75$ & \cellbg $85.97$ \\

\midrule
& \multicolumn{8}{c}{\texttt{\textbf{60s}}} \\
\cmidrule(lr){2-9}
\textbf{Method}
& \texttt{\textbf{Subj.}}$\uparrow$
& \texttt{\textbf{Backg.}}$\uparrow$
& \texttt{\textbf{Motion}}$\uparrow$
& \texttt{\textbf{Dyn.}}$\uparrow$
& \texttt{\textbf{Aesth.}}$\uparrow$
& \texttt{\textbf{Imaging}}$\uparrow$
& \texttt{\textbf{Flicker}}$\uparrow$
& \texttt{\textbf{Quality}}$\uparrow$ \\
\midrule

\rowcolor{gray!12}
\multicolumn{9}{l}{\textbf{\textit{Short-video supervision}}} \\

Self Forcing{\tiny~\citep{huang2025self}}
& $97.27$ & $96.33$ & $98.66$ & $44.21$ & $56.58$
& $67.87$ & $99.13$ & \cellgr $81.89$ \\

Causal Forcing{\tiny~\citep{zhu2026causal}}
& $95.96$ & $96.17$ & $97.52$ & $59.86$ & $51.83$
& $65.15$ & $98.77$ & \cellgr $80.93$ \\

LongLive{\tiny~\citep{yang2026longlive}}
& \cellte $98.40$ & \cellbg $97.10$ & \cellte $98.80$
& $37.04$ & $62.23$
& $69.48$ & \cellte $99.32$ & \cellgr $82.98$ \\

Reward Forcing{\tiny~\citep{lu2026reward}}
& $97.29$ & $96.13$ & $98.45$ & $80.23$ & $62.08$
& $69.95$ & $98.79$ & \cellgr $85.54$ \\

MemRoPE{\tiny~\citep{kim2026memrope}}
& $97.48$ & $96.33$ & $98.69$ & $56.20$ & $59.34$
& \cellbg $72.20$ & $99.03$ & \cellgr $83.92$ \\

Rolling Forcing$^{\dagger}${\tiny~\citep{liu2026rolling}}
& $95.81$ & $95.14$ & $97.63$ & \cellte $95.69$ & $56.75$
& $68.67$ & $98.72$ & \cellgr $84.78$ \\

Self Gradient Forcing{\tiny~\citep{zhuang2026self}}
& \cellbg $98.42$ & $96.47$ & $98.77$
& $53.06$ & \cellbg $64.09$
& $69.40$ & $98.87$ & \cellgr $84.15$ \\

\midrule
\rowcolor{gray!12}
\multicolumn{9}{l}{\textbf{\textit{Long-video supervision}}} \\

Context Forcing{\tiny~\citep{chen2026context}}
& $98.17$ & \cellte $96.75$ & \cellbg $99.00$
& $52.69$ & $62.10$
& $67.37$ & \cellbg $99.39$ & \cellgr $83.85$ \\

\rowcolor{gray!6}
\textbf{LongTake}{\scriptsize~\textbf{(Ours)}}
& $96.78$ & $95.52$  & $97.64$ & $94.72$ & \cellte $62.35$
& $69.91$ & $98.60$ & \cellbg $85.96$ \\

\rowcolor{gray!6}
\hspace{4mm}+ Hybrid DMD\gray{{\tiny~$(\lambda=0.2)$}}
& $96.43$ & $94.57$ & $97.47$ & \cellbg $97.36$ & $61.48$
& \cellte $70.17$ & $98.57$ & \cellte $85.71$ \\

\bottomrule
\end{tabular}%
}
\label{table:full_vbench_table}
\end{table*}

\subsection{Evaluation Protocol}
\label{appendix:evaluation-protocol}

\paragraph{Prompts and generation.}
The 30s evaluation uses the first 128 prompts from the extended MovieGen prompt set~\citep{polyak2024movie} distributed with Self Forcing~\citep{huang2025self}.
For 60s evaluation, we use the union of the six VBench-Long~\citep{huang2025vbench++} quality-dimension prompt sets, containing 251 unique prompts.
Within each evaluation horizon, all methods use the same prompts.
Table~\ref{tab:appendix-evaluation-prompts} lists the number of prompts evaluated for each video-quality dimension.

\begin{table}[t]
\centering
\caption{Evaluation prompt counts.}
\label{tab:appendix-evaluation-prompts}
\small
\begin{tabular}{lcc}
\toprule
Dimension & 30s & 60s \\
\midrule
Subject consistency & 128 & 72 \\
Background consistency & 128 & 86 \\
Motion smoothness & 128 & 72 \\
Dynamic degree & 128 & 72 \\
Aesthetic quality & 128 & 93 \\
Imaging quality & 128 & 93 \\
Temporal flickering candidates & 128 & 75 \\
\bottomrule
\end{tabular}
\end{table}

\paragraph{IQ Drift.}
Following~\citep{liu2026rolling}, we measure endpoint image-quality changes using the VBench MUSIQ evaluator~\citep{ke2021musiq} with the SPAQ checkpoint~\citep{fang2020perceptual}.
For each video, we average frame-level scores over the first and last five seconds, using all 80 frames in each window.
Let $m(\bx_{i,t})$ denote the MUSIQ score on its original point scale and let $W_{\ttb{start}}$ and $W_{\ttb{end}}$ denote these windows.
We report the mean absolute difference between the two endpoint scores as
\begin{equation}
\Delta\texttt{\textbf{IQ}}
= \frac{1}{N}\sum_{i=1}^{N}
\left|
\frac{1}{80}\sum_{t\in W_{\ttb{end}}}m(\bx_{i,t})
-\frac{1}{80}\sum_{t\in W_{\ttb{start}}}m(\bx_{i,t})
\right|.
\label{eq:appendix-iq-drift}
\end{equation}
We take the absolute difference before averaging over the 128 MovieGen videos at 30s or the 93-video imaging-quality subset at 60s.
Windows are extracted losslessly from the original RGB videos without further temporal splitting.
We use the longer preprocessing mode, which resizes $832\times480$ frames to $512\times295$ and scales pixel values by $1/255$.
Lower $\Delta\texttt{\textbf{IQ}}$ indicates smaller endpoint quality changes.
However, consistently low-quality videos and videos with intermediate degradation can also obtain low scores, so we interpret $\Delta\texttt{\textbf{IQ}}$ together with imaging quality.
To distinguish endpoint degradation from improvement, we additionally retain the signed end-minus-start difference, whose negative mean indicates degradation in endpoint image quality on average.

\paragraph{VLM Exposure.}
We use the exposure-stability rubric of MemRoPE~\citep{cui2026selfforcing, kim2026memrope} to evaluate overexposure, underexposure, and the resulting loss of visibility.
We use Gemini-3.1-pro-preview (model identifier $\ttb{gemini}$-$\ttb{3.1}$-$\ttb{pro}$-$\ttb{preview}$) as the judge.
Each request contains one full original 30s or 60s video
and the unchanged exposure rubric,
without the text prompt used for generation.
We fix the API settings to 1 FPS analysis sampling, high media resolution, temperature 1.0, low thinking level, and a maximum of 4,096 output tokens.
The original video remains at 16 FPS and is uploaded without re-encoding or temporal splitting.
The judge returns an integer score $e_i\in\{0,1,2,3,4,5\}$ and a textual justification.
Table~\ref{tab:appendix-exposure-rubric} summarizes the scoring levels, where higher scores indicate better exposure quality and stability.
We report the mean over all 128 MovieGen prompts~\citep{polyak2024movie}
for 30s generation and all 251 VBench-Long prompts~\citep{huang2024vbench} for 60s generation.
Failed requests are retried without assigning a score, while successful judgments are not repeated for score selection.
Each reported mean includes valid judgments for every video in its prompt set under the same unchanged rubric and fixed API settings.

\begin{table}[t]
\centering
\caption{\texttt{\textbf{VLM}} exposure-stability levels, summarized from the original MemRoPE rubric.}
\label{tab:appendix-exposure-rubric}
\small
\begin{tabular}{cp{0.82\linewidth}}
\toprule
Score & Exposure criterion \\
\midrule
0 & Near-total whiteout or blackout makes the scene unreadable. \\
1 & Widespread exposure failure severely reduces visibility. \\
2 & Persistent highlight or shadow clipping causes substantial detail loss. \\
3 & Exposure problems have limited spatial extent or duration. \\
4 & Occasional local exposure flaws cause little visibility loss. \\
5 & Balanced exposure preserves visibility without disruptive clipping or darkening. \\
\bottomrule
\end{tabular}
\end{table}

\paragraph{Aggregate score.}
We compute the Quality score using the standard VBench~\citep{huang2024vbench} normalization and weighting of the seven video-quality dimensions.
Dynamic degree receives a weight of 0.5, while each remaining dimension receives a weight of 1, and we scale the weighted mean by 100.
Separately, \textbf{\texttt{$\Delta$IQ}} and \textbf{\texttt{VLM}} Exposure assess visual stability and do not contribute to the Quality score.

\begin{figure*}[t]
\begin{minipage}{1.\linewidth}
    \includegraphics[width=1.\linewidth]{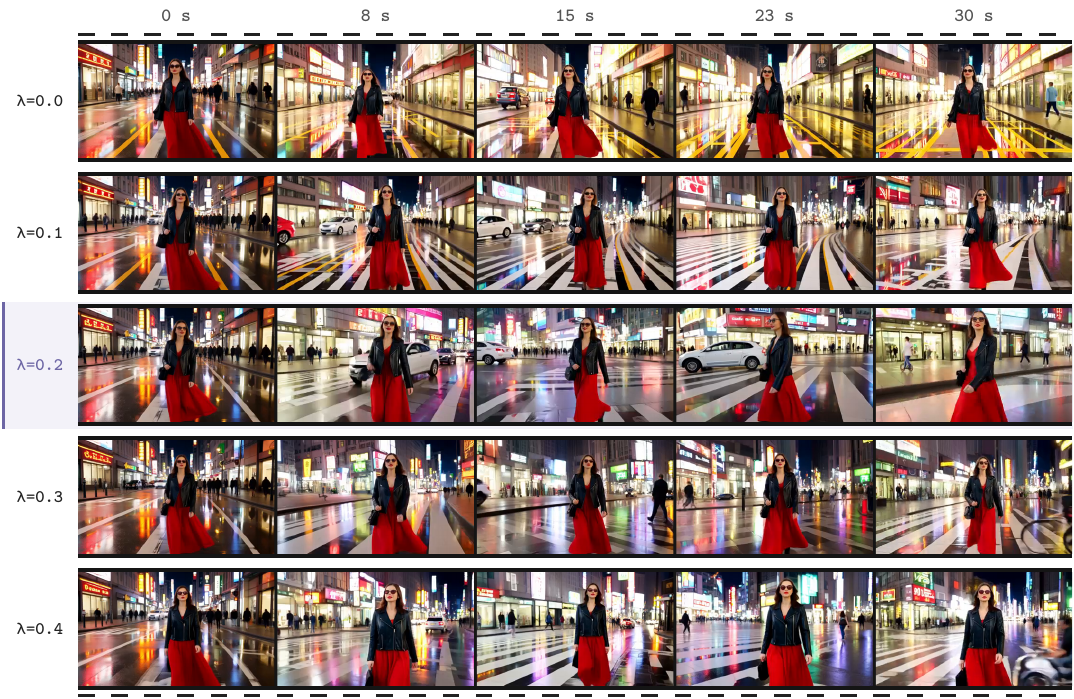}
\end{minipage}
\caption{
\textbf{Effect of the conditional DMD weight $\lambda$.}
Frames sampled from 30s rollouts
show how $\lambda$ affects scene progression.
With $\lambda=0.2$, both the viewpoint and surrounding scene evolve
while preserving a coherent appearance of the central subject throughout the rollout.
}
\label{fig:lambda_ablation}
\end{figure*}

\section{Additional Results}

\subsection{Teacher Generation Before Distillation}
\label{appendix:teacher-generation}

We compare Short TF and Long TF before DMD
by generating subsequent frames from the same real video prefix.
We use 50 held-out real long videos
that were not used for Long-Horizon TF fine-tuning.
For each video, each teacher constructs its conditioning KV-cache
from approximately 5s or 20s of real video
(81 or 321 video frames, respectively)
and then generates approximately 5s of additional video
(81 video frames), yielding 200 outputs in total.
Each teacher updates its conditioning KV-cache from generated frames,
retaining  3 sink frames and 15 recent latent frames.

\paragraph{Evaluation.}
We evaluate 24 uniformly sampled generated frames,
excluding the real video prefix.
VisionReward (\textbf{\texttt{VR}})~\citep{xu2024visionreward}
aggregates 29 binary video judgments using the official weights.
Instruction following (\textbf{\texttt{IF}})
evaluates whether the generated video satisfies
some requirements of the prompt.
Responses are encoded as $+1$ for yes and $-1$ for no
and averaged across videos.
Both scores are scaled by 100,
with \textbf{\texttt{IF}} reported as a signed average.
We assess the immediate transition from observed to generated frames
using boundary \textbf{\texttt{DINO}} feature similarity~\citep{caron2021emerging}
and normalized RGB L1 difference
between the last observed frame and the first generated frame.

\begin{table}[t]
\centering
\caption{
\textbf{Teacher generation before DMD on held-out real long videos.}
Each teacher constructs its conditioning KV-cache from the same real video prefix
and generates 81 additional video frames.
Quality and prompt alignment are evaluated on generated frames,
while boundary metrics assess appearance preservation
at the immediate transition from observed to generated video frames.
}
\label{tab:teacher-generation}
\small
\setlength{\tabcolsep}{5pt}
\begin{tabular}{@{}llrrrr@{}}
\toprule
Prefix (video frames) & Teacher
& \textbf{\texttt{VR}} $\uparrow$
& \textbf{\texttt{IF}} $\uparrow$
& \textbf{\texttt{DINO}} $\uparrow$
& RGB L1 $\downarrow$ \\
\midrule
81 $(\approx 5s)$ & Short TF & 17.26 & 96 & 0.9904 & 0.0202 \\
& Long TF & \cellbg 18.28 & \cellbg 100
& \cellbg 0.9919 & \cellbg 0.0176 \\
\midrule
321$(\approx 20s)$ & Short TF & 16.60 & 96 & 0.9930 & 0.0183 \\
& Long TF & \cellbg 17.83 & \cellbg 100
& \cellbg 0.9941 & \cellbg 0.0166 \\
\bottomrule
\end{tabular}
\end{table}

\paragraph{Results.}
In~\Cref{tab:teacher-generation}, the Long-Horizon TF model trained on curated real long videos yields improvements in evaluated generation quality at both prefix lengths.
The boundary metrics suggest better preservation of visual appearance
at the transition from observed to generated frames,
although these measurements are limited to the immediate transition
and can favor static outputs.
Because \textbf{\texttt{IF}} scores are near saturation,
they provide limited evidence of a difference
in prompt alignment between the two teachers.

\begin{figure*}[!t]
\begin{minipage}{1.\linewidth}
    \includegraphics[width=1.\linewidth]{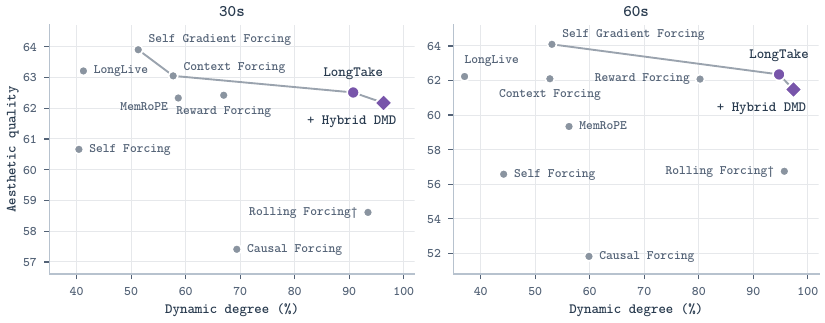}
\end{minipage}
\caption{
\textbf{Motion-quality trade-off at (Left) 30s and (Right) 60s.}
The gray line joins the Pareto-optimal methods. No other method has both higher dynamic degree and higher aesthetic quality. Both LongTake variants lie on front at both durations.
}
\label{fig:motion-quality trade-off_appen1}
\end{figure*}

\begin{figure*}[!t]
\begin{minipage}{1.\linewidth}
    \includegraphics[width=1.\linewidth]{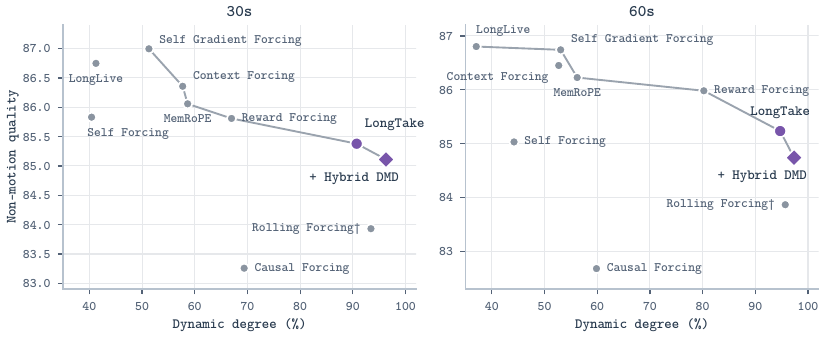}
\end{minipage}
\caption{
\textbf{Motion vs. non-motion quality}
Non-motion quality is the VBench Quality score recomputed without dynamic degree with equal weights. Both LongTake variants lie on the front at both durations and are the only methods on it with dynamic degree above $90$.
}
\label{fig:motion-quality trade-off_appen2}
\end{figure*}

\subsection{Motion-Quality Trade-off}
\label{appendix:motion-quality trade-off}

\Cref{fig:motion-quality trade-off_appen1} extends~\Cref{fig:dynamics_aesthetics} to $60$s rollouts. Both LongTake variants remain on the Pareto front of dynamic degree and aesthetic quality, and the same holds when aesthetic quality is replaced by non-motion quality, the VBench Quality score recomputed without dynamic degree (\Cref{fig:motion-quality trade-off_appen2}).

\subsection{Additional Qualitative Results}
\label{appendix:qualitative-results}

We complement the quantitative evaluation with additional visual comparisons of long-horizon generation, TF initialization, and conditional supervision through Hybrid DMD.

\paragraph{Effect of the conditional DMD weight.}
\Cref{fig:lambda_ablation}
compares 30s rollouts with different values of $\lambda$.
In this example, $\lambda=0.2$
produces more pronounced scene progression
than $\lambda=0$,
while preserving a coherent appearance of the subject throughout the generated sequence.

\begin{figure}[t]
    \centering
    \includegraphics[width=\linewidth]{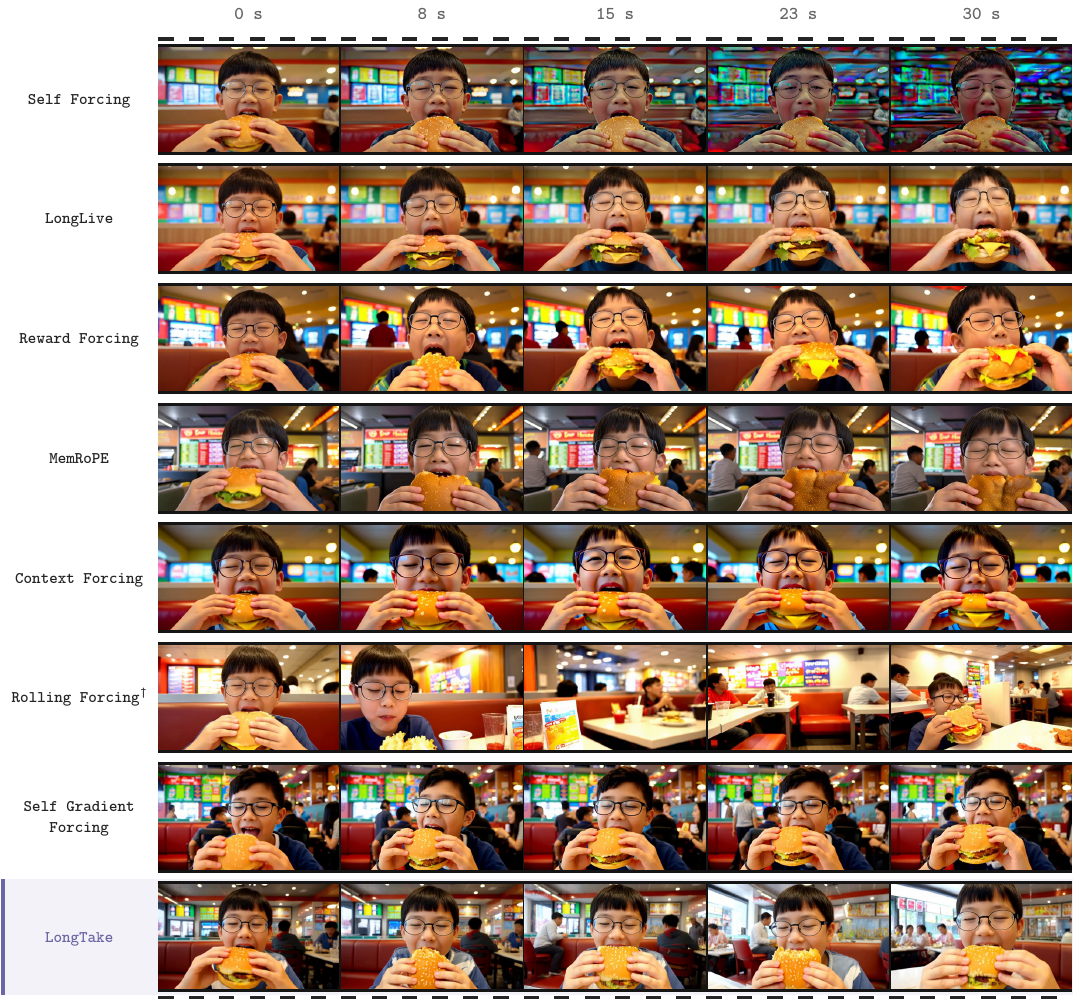}
    \caption{\textbf{30s generation with the same prompt across evaluated methods.}}
    \label{fig:long_generation_1}
\end{figure}

\begin{figure}[t]
    \centering
    \includegraphics[width=\linewidth]{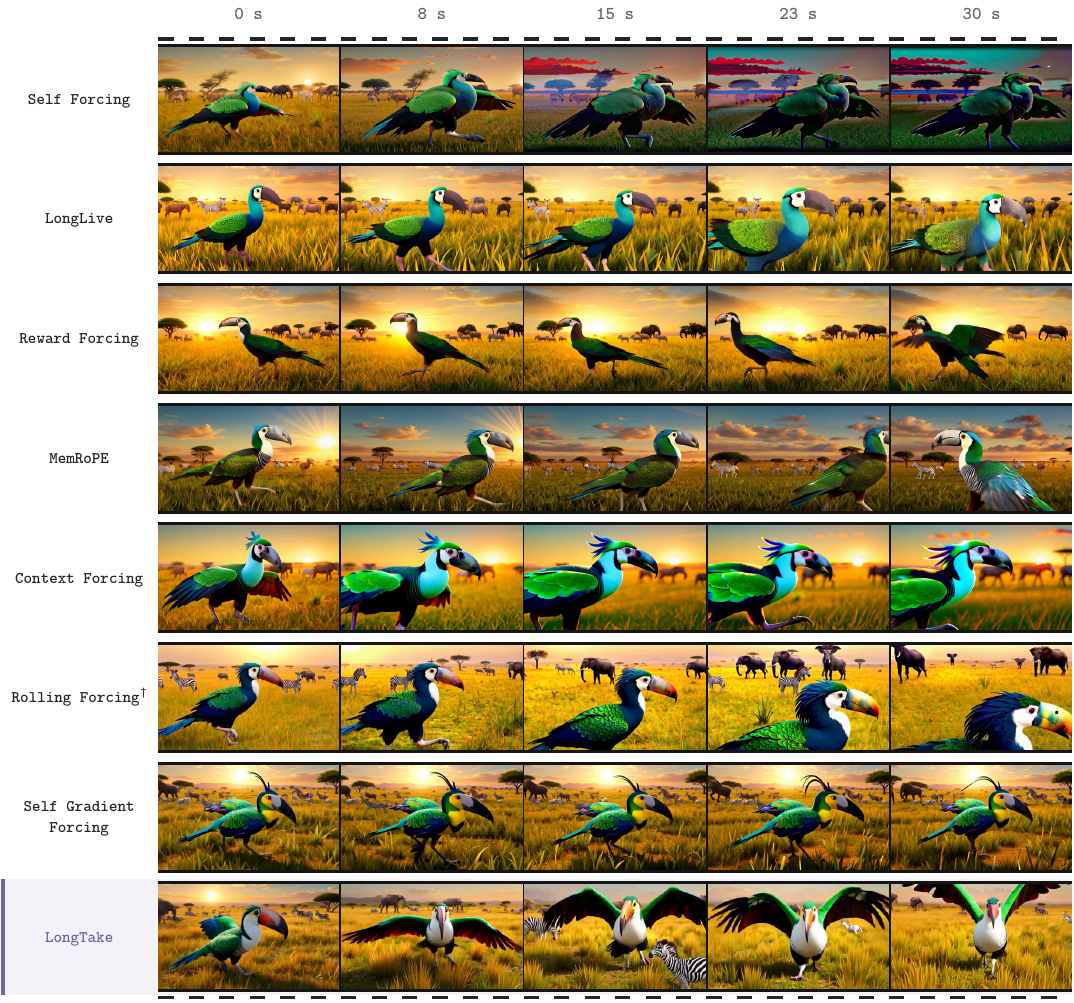}
    \caption{\textbf{30s generation with the same prompt across evaluated methods.}}
    \label{fig:long_generation_2}
\end{figure}

\begin{figure}[t]
    \centering
    \includegraphics[width=\linewidth]{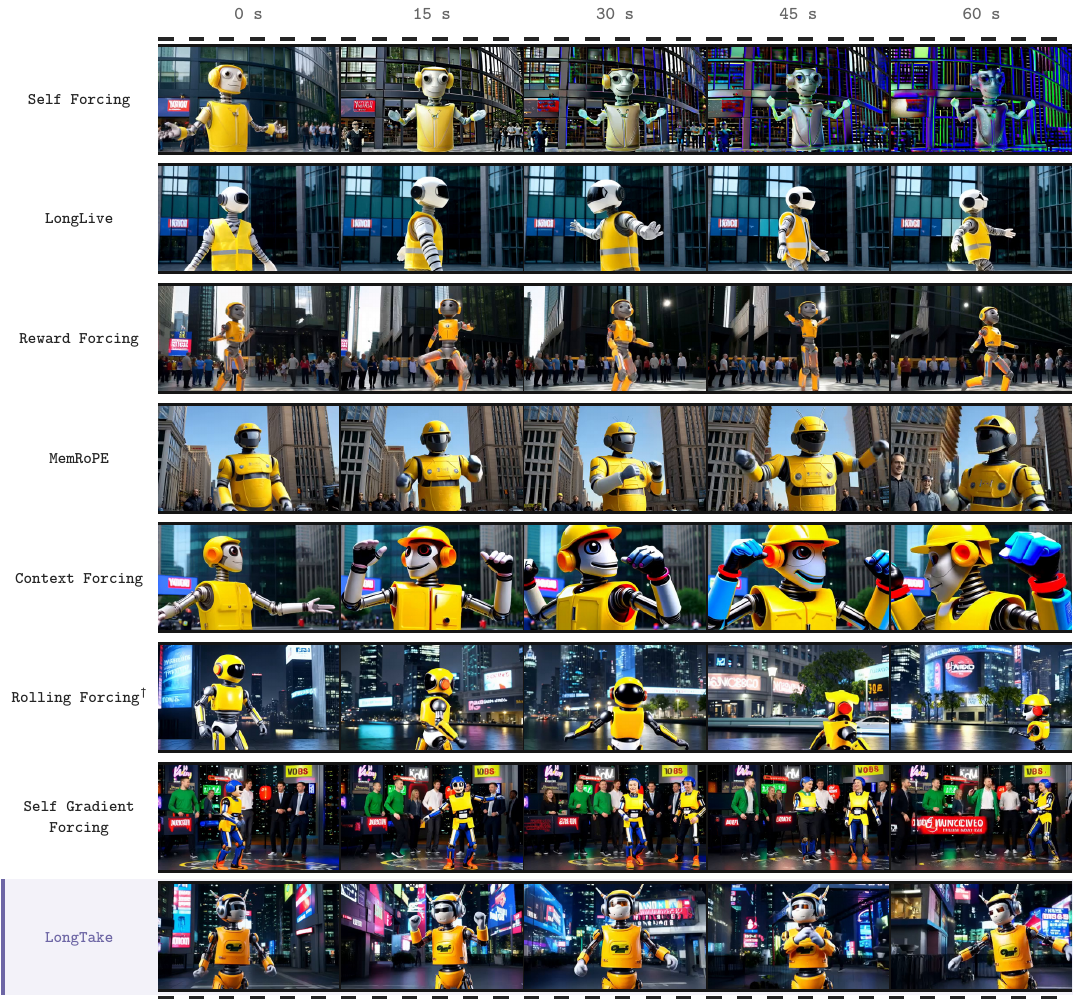}
    \caption{\textbf{60s generation with the same prompt across evaluated methods.}}
    \label{fig:long_generation_3}
\end{figure}

\paragraph{Long-horizon generation.}
\Cref{fig:long_generation_1,fig:long_generation_2,fig:long_generation_3}
provide additional baseline comparisons on 30s and 60s generation, complementing the examples in~\Cref{fig:qualitative_30s_60s} with different subjects and scenes.

\begin{figure}[t]
    \centering
    \includegraphics[width=\linewidth]{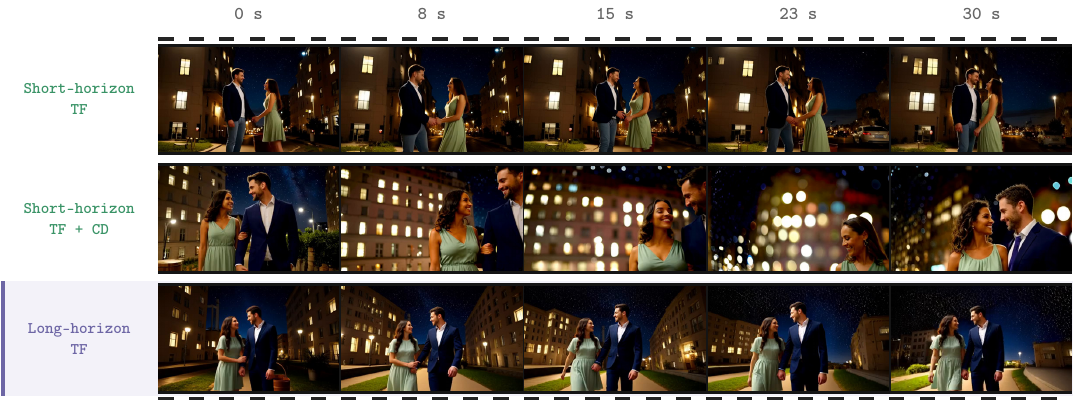}
    \caption{\textbf{Long-Horizon TF initialization under the same 5s joint DMD.}}
    \label{fig:long_tf_1}
\end{figure}

\begin{figure}[t]
    \centering
    \includegraphics[width=\linewidth]{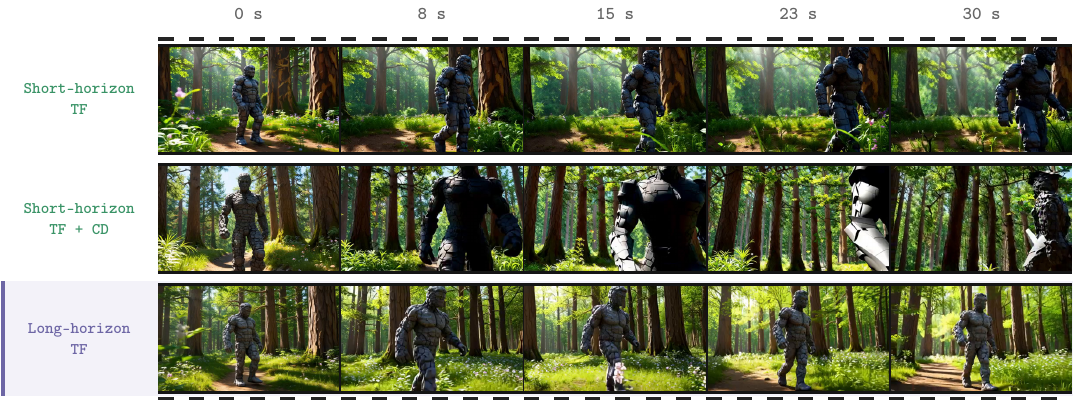}
    \caption{\textbf{Long-Horizon TF initialization under the same 5s joint DMD.}}
    \label{fig:long_tf_2}
\end{figure}

\begin{figure}[t]
    \centering
    \includegraphics[width=\linewidth]{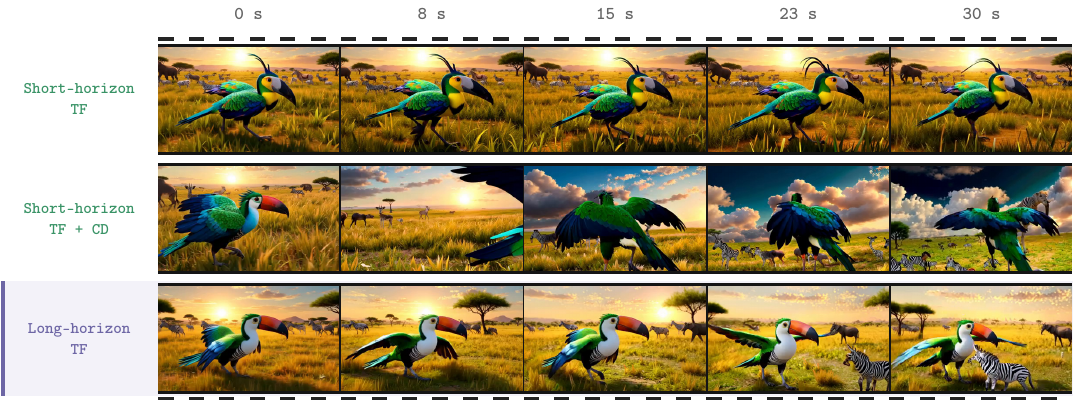}
    \caption{\textbf{Long-Horizon TF initialization under the same 5s joint DMD.}}
    \label{fig:long_tf_3}
\end{figure}

\paragraph{Effect of Long-Horizon TF.}
\Cref{fig:long_tf_1,fig:long_tf_2,fig:long_tf_3}
provide additional comparisons
illustrating the effect of Long-Horizon TF initialization under the same 5s joint DMD procedure.

\begin{figure}[t]
    \centering
    \includegraphics[width=\linewidth]{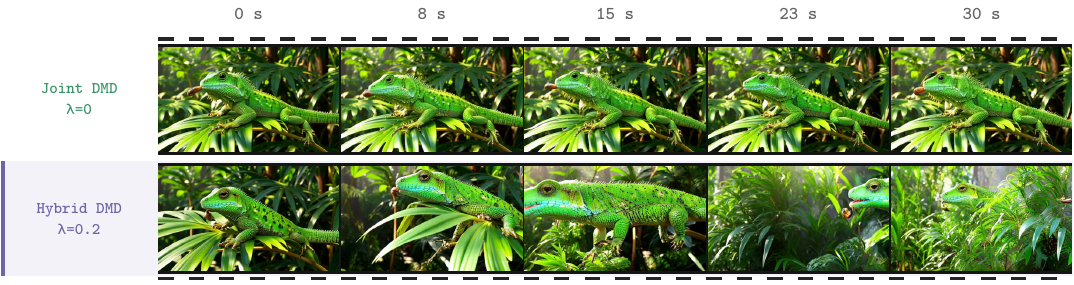}
    \caption{\textbf{Hybrid DMD with shared initialization and equal distillation iterations.}}
    \label{fig:hybrid_dmd_1}
\end{figure}

\begin{figure}[t]
    \centering
    \includegraphics[width=\linewidth]{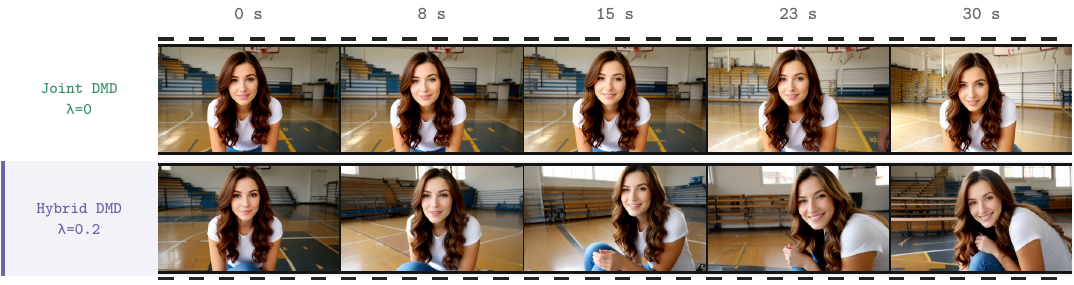}
    \caption{\textbf{Hybrid DMD with shared initialization and equal distillation iterations.}}
    \label{fig:hybrid_dmd_2}
\end{figure}

\paragraph{Effect of Hybrid DMD.}
\Cref{fig:hybrid_dmd_1,fig:hybrid_dmd_2}
provide additional comparisons
illustrating the effect of Hybrid DMD.
These examples complement
Table~\ref{tab:lambda_ablation},
where all settings use the same Long-Horizon TF initialization
and are evaluated after 1,200 distillation iterations.
Compared with joint DMD ($\lambda=0$), Hybrid DMD ($\lambda=0.2$)
increases dynamics,
with modest reductions in subject consistency and aesthetic quality.
With the teacher-training stage held fixed, this comparison shows the additional motion--quality tradeoff introduced by extending supervision through Hybrid DMD.

\end{document}

%% file: math.tex
\usepackage{cleveref}
\usepackage{xcolor,soul}
\usepackage{colortbl}

\newcommand{\bx}{\mathbf{x}}

\newcommand{\calC}{{\mathcal{C}}}
\newcommand{\calD}{{\mathcal{D}}}

\newcommand{\calJ}{{\mathcal{J}}}

\newcommand{\calN}{{\mathcal{N}}}

\newcommand{\bbE}{\mathbb{E}}

\def\[#1\]{\begin{align}#1\end{align}}
\newcommand{\norm}[1]{\left\lVert{#1}\right\rVert}

\newcommand{\ie}{\textit{i}.\textit{e}., }
\newcommand{\eg}{\textit{e}.\textit{g}., }

\theoremstyle{plain}

\def\[#1\]{\begin{align}#1\end{align}}

\newcommand{\gray}[1]{{\color{gray}{#1}}}

\definecolor{myellow}{RGB}{194, 125, 47}
\definecolor{mgreen}{RGB}{48, 160, 111}

\definecolor{mteal}{RGB}{230,250,249}
\definecolor{bgteal}{RGB}{236, 245, 245}
\definecolor{mpurple}{RGB}{120, 111, 177}
\definecolor{citationcolor}{RGB}{80, 90, 180}

\newcommand{\green}[1]{{\color{mgreen}{#1}}}

\newcommand{\purple}[1]{{\color{mpurple}{#1}}}

\usepackage{thmtools}
\usepackage{thm-restate}

\definecolor{mygray}{gray}{0.95}
\newcommand{\graybox}[1]{%
\begingroup
\setlength{\fboxsep}{0pt}%
\colorbox{mygray} {
\begin{minipage}{\linewidth}
\vspace{-0.5em}%
{#1}%
\end{minipage}%
}
\endgroup
}

\definecolor{mmgreen}{RGB}{243, 247, 243}
\definecolor{mmpurple}{RGB}{246, 242, 247}
\definecolor{mmccolor}{RGB}{235, 242, 250}

\definecolor{citationcolor}{RGB}{80, 90, 180}
\definecolor{background}{RGB}{215, 215, 235}
\definecolor{bggreen}{RGB}{229,242,229}
\newcommand{\cellbg}{\cellcolor{background}}
\newcommand{\cellte}{\cellcolor{mteal}}
\newcommand{\cellgr}{\cellcolor{gray!4}}

\DeclareRobustCommand{\best}[1]{{\sethlcolor{background}\hl{#1}}}
\DeclareRobustCommand{\secondbest}[1]{{\sethlcolor{mteal}\hl{#1}}}

\newsavebox\CBox

%% file: math_commands.tex
\usepackage{amsmath,amsfonts,bm}

\def\eqref#1{(\ref{#1})}

\def\1{\bm{1}}

\DeclareMathAlphabet{\mathsfit}{\encodingdefault}{\sfdefault}{m}{sl}
\SetMathAlphabet{\mathsfit}{bold}{\encodingdefault}{\sfdefault}{bx}{n}

\newcommand{\KL}{D_{\mathrm{KL}}}

\DeclareMathOperator*{\argmin}{arg\,min}